\documentclass[10pt,twocolumn,letterpaper]{article}
\pdfoutput=1
\usepackage{cvpr}
\usepackage[accsupp]{axessibility}
\usepackage{graphicx}
\usepackage{amsmath}
\usepackage{amssymb}
\usepackage{booktabs}
\usepackage{animate}
\usepackage{enumitem}
\usepackage{multirow}
\usepackage{diagbox}
\usepackage[table,xcdraw]{xcolor}
\usepackage{float}
\usepackage{diagbox}
\usepackage{tabularx}
\usepackage{microtype}
\usepackage{soul}

\makeatletter
\@namedef{ver@everyshi.sty}{}
\makeatother
\usepackage{tikz}
\usepackage{makecell}

\definecolor{cvprblue}{rgb}{0.21,0.49,0.74}
\usepackage[pagebackref,breaklinks,colorlinks,allcolors=cvprblue]{hyperref}

\usepackage[capitalize]{cleveref}

\definecolor{Gray}{gray}{0.5}
\definecolor{LightCyan}{rgb}{0.88,1,1}

\newcolumntype{a}{>{\columncolor{Gray}}c}
\newcolumntype{b}{>{\columncolor{white}}c}

\def\ie{\textit{i.e.}}
\def\eg{\textit{e.g.}}

\definecolor{top1}{RGB}{255,179,179}
\definecolor{top2}{RGB}{255,217,179}
\definecolor{top3}{RGB}{255,255,179}

\begin{document}

%%%%%%%%% TITLE - PLEASE UPDATE

\title{NexusGS: Sparse View Synthesis with Epipolar Depth Priors \\in 3D Gaussian Splatting}

\author{Yulong Zheng$^1$, Zicheng Jiang$^1$, Shengfeng He$^2$, Yandu Sun$^1$, Junyu Dong$^1$, Huaidong Zhang$^3$, Yong Du$^1$\thanks{Corresponding author (csyongdu@ouc.edu.cn).}\\
	$^1$ Ocean University of China,\\
	$^2$ Singapore Management University, $^3$ South China University of Technology
}

\teaser{
    \centering
      
    \includegraphics[width=\linewidth]{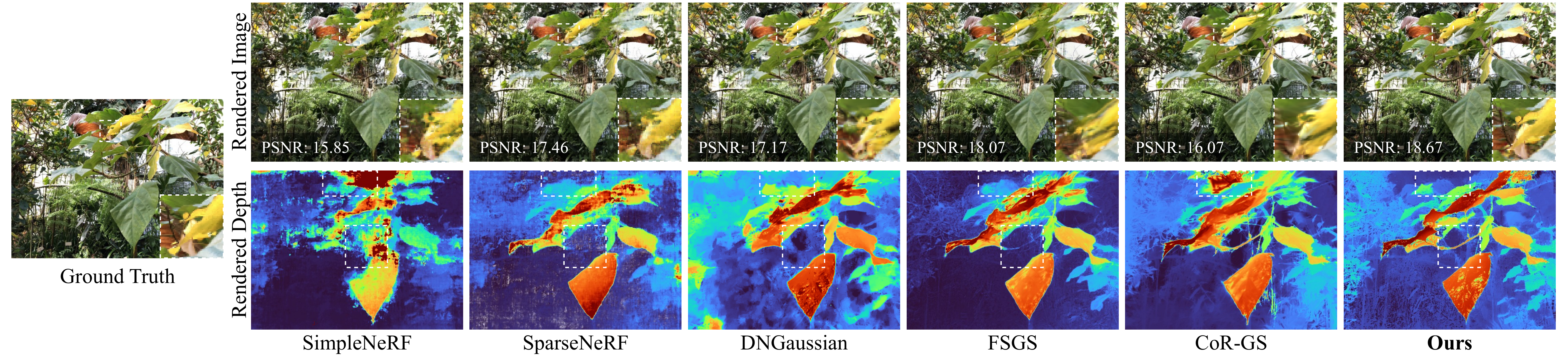}
    \caption{NexusGS distinguishes itself from both NeRF-based and 3DGS-based competitors by incorporating epipolar depth priors, significantly improving the accuracy of depth maps and enhancing the fidelity of rendered images. This effectiveness in handling sparse input views is achieved through innovative point cloud densification with depth blending and pruning strategies.}\vspace{-3mm}
    \label{fig:teaser}\vspace{1mm}
}
\maketitle

\begin{abstract}
Neural Radiance Field (NeRF) and 3D Gaussian Splatting (3DGS) have noticeably advanced photo-realistic novel view synthesis using images from densely spaced camera viewpoints. However, these methods struggle in few-shot scenarios due to limited supervision. In this paper, we present NexusGS, a 3DGS-based approach that enhances novel view synthesis from sparse-view images by directly embedding depth information into point clouds, without relying on complex manual regularizations. Exploiting the inherent epipolar geometry of 3DGS, our method introduces a novel point cloud densification strategy that initializes 3DGS with a dense point cloud, reducing randomness in point placement while preventing over-smoothing and overfitting. Specifically, NexusGS comprises three key steps: Epipolar Depth Nexus, Flow-Resilient Depth Blending, and Flow-Filtered Depth Pruning. These steps leverage optical flow and camera poses to compute accurate depth maps, while mitigating the inaccuracies often associated with optical flow. By incorporating epipolar depth priors, NexusGS ensures reliable dense point cloud coverage and supports stable 3DGS training under sparse-view conditions. Experiments demonstrate that NexusGS significantly enhances depth accuracy and rendering quality, surpassing state-of-the-art methods by a considerable margin. Furthermore, we validate the superiority of our generated point clouds by substantially boosting the performance of competing methods. Project page: \url{https://usmizuki.github.io/NexusGS/}.
\end{abstract}

\section{Introduction}
Neural Radiance Field (NeRF)~\cite{mildenhall2021nerf} and 3D Gaussian Splatting (3DGS)~\cite{kerbl20233d} have recently advanced the field of photo-realistic novel view synthesis by leveraging images captured from densely spaced camera viewpoints to reconstruct 3D scenes under supervised conditions. However, these methods exhibit limitations in few-shot scenarios due to the reduced availability of supervision. Consequently, recent research has focused on improving the generalization of NeRF and 3DGS models in sparse-view settings to enhance their applicability in real-world contexts such as virtual reality and augmented reality.

Sparse view synthesis methods based on NeRF~\cite{yang2023freenerf,wang2023sparsenerf} typically rely on backpropagation combined with additional priors and regularization strategies to guide the learning of implicit radiance fields. Despite their progress, they inherit a key limitation of NeRF: the need for random sampling during volumetric rendering along each ray, which results in high training and rendering costs. While subsequent acceleration techniques~\cite{barron2021mip,barron2022mip,yu2021plenoctrees,sun2022direct} face a trade-off between computational expense and rendering quality.

In contrast, 3DGS uses Gaussian primitives to model 3D geometry from sparse point clouds, typically generated by Structure-from-Motion (SfM)~\cite{schonberger2016structure}, and offers a more efficient solution with its splat-based rasterization, enabling real-time, high-quality rendering. However, the application of 3DGS in sparse view synthesis is less explored compared to NeRF-based methods. A major challenge lies in densifying the initial point clouds to ensure comprehensive scene coverage with limited data, which depends critically on two factors: the number of points and their positional accuracy.

In 3DGS, the number of points is increased through a split operation during adaptive density control. In dense-view novel view synthesis, rich supervision effectively reduces randomness in point placement. However, this method struggles in sparse-view scenarios due to the lack of constraints, leading to oversmoothed reconstructions and poor generalization. FSGS~\cite{zhu2023fsgs} addresses this by introducing a Gaussian Unpooling mechanism, which generates new Gaussians at the centers of edges connecting a prototype Gaussian to its $K$-nearest neighbors. Nevertheless, this approach often results in point clustering, neglecting areas between clusters and hindering the formation of a reliable dense point cloud.

On the other hand, improving the positional accuracy of 3D points often requires incorporating additional geometric priors. Depth information serves as a crucial link between 2D imagery and 3D reconstruction, with many methods~\cite{wang2023sparsenerf,li2024dngaussian,zhu2023fsgs} leveraging depth maps predicted by monocular estimators~\cite{ranftl2021vision,ranftl2020towards,bhat2023zoedepth}. These approaches typically impose depth-aware regularizations to address the scale ambiguity inherent in monocular perception. However, these handcrafted constraints often prove inadequate, as they lack explicit metric grounding and merely redistribute rather than eliminate scale errors. 

In this paper, we introduce NexusGS, a 3DGS-based method for novel view synthesis from limited-view images. Our approach embeds depth information into point clouds to leverage the epipolar representation capabilities of 3DGS, enabling accurate geometric modeling without the need for complex manual regularization. The core of our method is a point cloud densification technique. Unlike traditional methods that gradually densify point clouds through split operations, our approach aims to form a relatively dense point cloud at the initial stage, reducing placement randomness and alleviating oversmoothing or overfitting in sparse view synthesis. This technique comprises three key steps: Epipolar Depth Nexus, Flow-Resilient Depth Blending, and Flow-Filtered Depth Pruning.

Drawing inspiration from triangulation, the Epipolar Depth Nexus step uses optical flow and camera poses within an epipolar geometry framework to compute depth accurately and explicitly. Given that optical flow estimates can be inaccurate, leading to inconsistent depth maps, we propose Flow-Resilient Depth Blending, which selects depths with minimal sensitivity to flow errors by evaluating the relative rate of change between reference and projection distances. Flow-Filtered Depth Pruning further refines the depth maps by removing points that are too far from the epipolar line, thus mitigating the impact of optical flow inaccuracies.

Although optical flow predictions may be imprecise, they tend to be more stable than depth maps due to their image-resolution-bound solution space. In contrast, depth maps span an unbounded domain, with errors propagating through scene-dependent scales, causing error bounds to grow linearly with the depth range. By leveraging optical flow for epipolar depth computation and incorporating our proposed strategies, NexusGS achieves reliable initial dense point cloud coverage. This foundation supports stable 3DGS training under sparse-view conditions, resulting in highly accurate depth maps and improved image quality (see Fig.~\ref{fig:teaser}).

In summary, the contributions of this work are threefold:
\begin{itemize}
\item We enhance sparse view synthesis by embedding epipolar depth priors into 3DGS, avoiding errors from monocular depth estimations and bypassing complex regularizations.
\item We present a point cloud densification technique for 3DGS that ensures accurate depth computation from sparse inputs using optical flow and camera poses, mitigates the impact of flow prediction inaccuracies, and delivers reliable dense point cloud coverage.
\item Extensive experiments demonstrate that NexusGS outperforms state-of-the-art methods in both depth accuracy and rendering quality across a variety of scenes, including real-world and object-centric scenarios. Furthermore, the superiority of our generated point clouds is validated by significantly enhancing the performance of competing methods.
\end{itemize}

\section{Related Work}
\textbf{Radiance Fields.}
Neural Radiance Fields (NeRFs) have revolutionized 3D scene reconstruction and novel view synthesis by utilizing neural representations via volumetric rendering. While NeRF produces high-quality images, its computational intensity leads to slow training and inference times. To address this, various techniques have been developed to improve efficiency~\cite{sun2022direct,reiser2021kilonerf,muller2022instant,garbin2021fastnerf}, along with methods focused on enhancing image quality~\cite{verbin2022ref,suhail2022light,guo2022nerfren,chen2022aug} and generalization capabilities~\cite{yu2021pixelnerf,wang2021ibrnet,wang2022attention,johari2022geonerf}. Complementing these efforts, 3D Gaussian Splatting (3DGS)~\cite{kerbl20233d,yu2024mip,cheng2024gaussianpro,huang20242d} offers an explicit alternative, using anisotropic 3D Gaussians for fast, high-quality, real-time rendering. As an explicit method, 3DGS excels in capturing fine details in complex scenes and continues to emerge as a robust approach alongside implicit methods like NeRF.\\
\textbf{Sparse View Synthesis Using NeRF.}
Synthesizing novel views from sparse input images is a critical yet challenging task in computer vision. Traditional NeRF models~\cite{barron2021mip,xu2022point}, which heavily rely on extensive data, struggle in sparse-view scenarios due to their dependence on dense input for accurate reconstruction. Recent research~\cite{yu2021pixelnerf,yang2023freenerf,zhu2024vanilla,shi2024zerorf} has sought to mitigate these challenges by incorporating various regularization techniques, such as depth supervision to improve training efficiency and reduce the number of required views~\cite{gao2022nerf}, geometry regularization to stabilize the learning process~\cite{niemeyer2022regnerf,somraj2023vip}, and semantic constraints using high-dimensional spaces like CLIP~\cite{radford2021learning,jain2021putting,gao2024hg3}.

As a core element of sparse-view synthesis, acquiring and leveraging depth priors effectively has become a key focus in the field. For example, early work such as DS-NeRF~\cite{deng2022depth} introduces a sparse depth prior to constrain the scene reconstruction process, while DDP-NeRF~\cite{roessle2022dense} further enriches this prior by explicitly modeling uncertainty, making it more comprehensive but less reliable. Simple-RF~\cite{somraj2024simple} proposes filtering out depth priors using image self-information but fails to obtain a dense depth prior without relying on reliable depth maps. In contrast, SCADE~\cite{uy2023scade} combines monocular depth estimates to obtain a better depth prior, yet this prior is learned through probabilistic optimization, making it uncertain and still sparse or unreliable.

In summary, most of these methods can only acquire sparse depth priors or unreliable dense priors due to inadequately suited regularizations. Moreover, NeRF's reliance on volume rendering continues to pose challenges, particularly in balancing rendering speed with image quality.

\begin{figure*}[htp]
	\centering
	\includegraphics[width=1\textwidth]{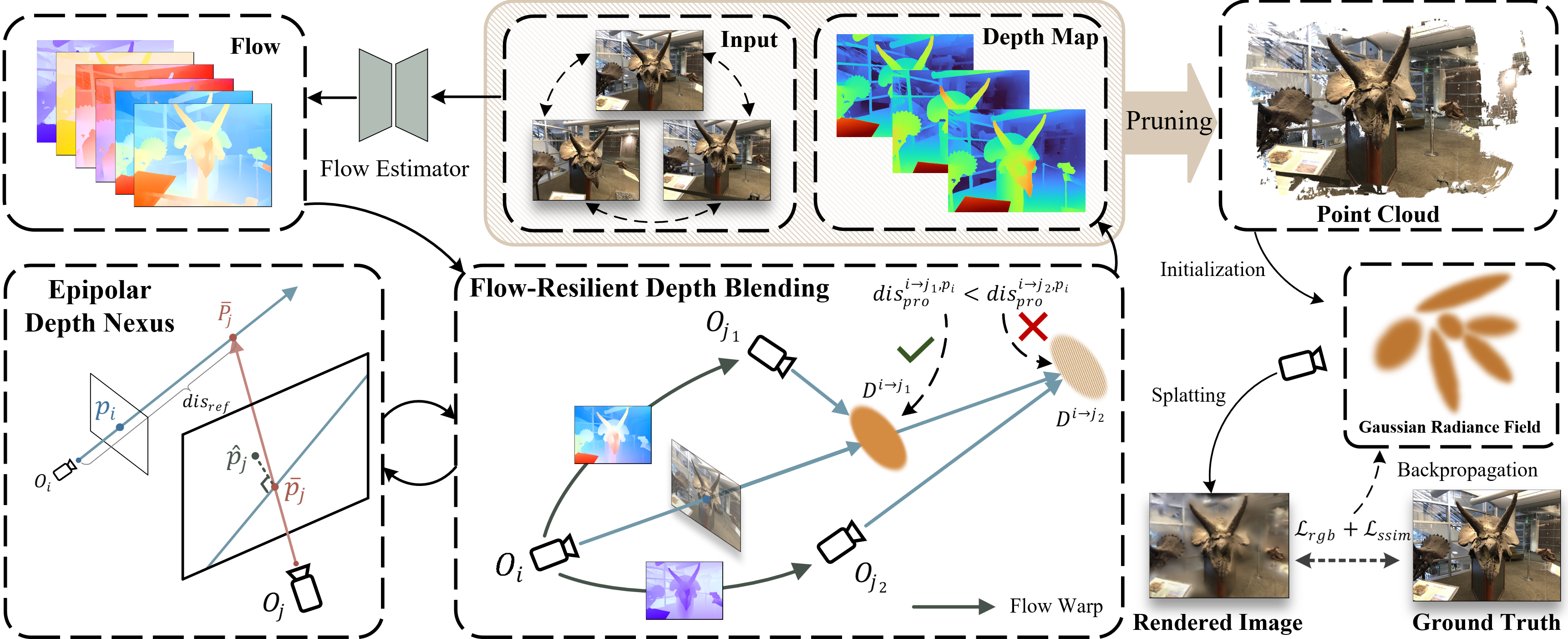}
	\caption{Given a few input images, our method first computes depth using optical flow and camera poses at the Epipolar Depth Nexus step. We then fuse depth values from different views, minimizing flow errors with flow-resilient depth blending. Before forming the final dense point cloud, outlier depths are removed at the flow-filter depth pruning step. In training, we do not need depth regularization, thanks to the embedded epipolar depth prior in the point cloud. }\vspace{-4mm}
	\label{fig:pipline_figure}
\end{figure*}

\noindent\textbf{Sparse View Synthesis Using 3DGS.} In contrast to the flourishing development of NeRF-based sparse view synthesis methods, the application of 3D Gaussian Splatting (3DGS) in sparse-view scenarios has encountered difficulties in maintaining consistent reconstructions with minimal input. Emerging methods, such as DR-3DGS~\cite{chung2024depth}, and FSGS~\cite{zhu2023fsgs}, attempt to address these issues through depth-based constraints or point cloud densification techniques. CoR-GS~\cite{zhang2024cor} trains and jointly refines multiple 3DGS fields simultaneously, constraining point cloud generation. CoherentGS~\cite{paliwal2025coherentgs} enforces pixel-Gaussian binding to suppress densification noise at the cost of constrained primitive counts. Despite these efforts, these approaches often suffer from inaccuracies arising from depth estimations, failing to adequately model the relationship between monocular depth estimates and true depth, underscoring the necessity for more robust strategies to improve 3DGS performance in sparse-view contexts. Our work builds on these foundations by proposing a novel co-regularization framework that overcomes the limitations of existing methods. Specifically, we focus on enhancing the consistency of Gaussian radiance fields in sparse-view scenarios using epipolar depth priors, moving beyond the reliance on monocular depth maps, avoiding the manual design of complex regularizations, and pushing the boundaries of sparse-view novel view synthesis.

We also recognize an additional class of sparse-view synthesis methods based on multi-view stereo (MVS), such as MVSNeRF~\cite{chen2021mvsnerf}, PixelSplat~\cite{charatan2024pixelsplat}, and MVSGaussian~\cite{liu2025mvsgaussian}. However, these methods rely on large datasets and require dense image inputs from the same scene for training to achieve high reconstruction quality, which falls outside the scope of this paper. In contrast, our approach operates within the constraints of sparse training images and does not require extensive pre-training, enabling us to attain high-quality reconstructions with minimal data requirements.
\section{Preliminaries}
\textbf{Gaussian Splatting.}
3D Gaussian Splatting~\cite{kerbl20233d} depicts 3D scenes using a set of Gaussians $G(\cdot)$, each defined as:
\begin{equation}
    G(x) = e^{-\frac{1}{2}(x-\mu)^\top\Sigma^{-1}(x-\mu)},
\end{equation}
where $x$ denotes a point in 3D space, and $\mu$ represent the center of the 3D Gaussian. The covariance matrix $\Sigma$ can be decomposed as $\Sigma = RSS^{\top} R^{\top}$, where $R \in \mathbb{R}^{3\times 3}$ is a rotation matrix, and $S \in \mathbb{R}^{3\times 1}$ are the scale parameters. Each Gaussian also includes spherical harmonics (SH) coefficients for color $c \in \mathbb{R}^{3\times 1}$, and an opacity value.

For pixel rendering, 3DGS sorts the 3D Gaussian primitives intersecting a pixel's camera ray. Let $Q$ be the ordered set of Gaussians along the ray, with $i$ representing the index of each element in $Q$. The renderer computes the color $c_i$ and opacity $\alpha_i$ for each Gaussian, \ie, $\{\alpha_i,c_i\mid i \in Q \}$, and determines the pixel color as follows: 
\begin{equation}
    C = \sum_{i \in Q}c_i \alpha_i \prod_{j=1}^{i-1}(1-\alpha_j),
    \label{eq:render}
\end{equation}
where the summation represents the accumulated color, and the product accounts for the cumulative transmittance.

During training, 3DGS initializes with a sparse point cloud and uses L1 and D-SSIM losses between the rendered and ground truth images to guide the optimization process.\\
\textbf{Epipolar Line.}
As shown on the left side of Fig.~\ref{fig:pipline_figure}, for a point $p_i$ in the source view $i$, the corresponding epipolar line $l_{p_i}$ in the target view $j$ can be determined as follows:
\begin{equation}
(a,b,c)^{\top} = F^{i\rightarrow j} (x_i, y_i, 1)^{\top},
\label{eq:compute_epipolar}
\end{equation}
where $F^{i\rightarrow j}$ is the fundamental matrix from the source camera to the target camera, and $(x_i,y_i)$ represents the coordinate of point $p_i$ in the source image. The epipolar line equation in the target camera's pixel coordinate system is then given by:
\begin{equation}
ax + by + c = 0.
\label{eq:epipolar_line}
\end{equation}
\section{Approach}
\textbf{Epipolar Depth Nexus.}
\label{sec:depth_computation}
A major challenge in 3DGS when applied to sparse-view novel view synthesis arises from the randomness introduced by the split operation during adaptive density control, combined with the limited supervision from sparse training data. To overcome this, NexusGS proposes initializing with a dense point cloud, as shown in the pipeline in Fig.~\ref{fig:pipline_figure}. 

Each point in the point cloud used for 3DGS has at least two key attributes: color and position. While input images can guide the color assignment, they cannot provide accurate depth values, which are crucial for determining the precise positions of the points. Monocular depth estimation networks often fail to deliver reliable depth information. In contrast, our approach overcomes this by establishing a nexus between the epipolar line and depth, using optical flow and camera parameters to compute accurate depth values.

To begin with, we predict optical flow between views using existing optical flow estimators~\cite{shi2023flowformer++} as follows:
\begin{equation}
    M_{flow}^{i\rightarrow j} = f(I_i, I_j),
    \label{eq:compute_flow}
\end{equation}
where $f(\cdot)$ denotes the pre-trained flow estimator, $I_i$ and $I_j$ are the given images of the source and target views, respectively. For a point $p_i$ in the source view, the predicted matched point $\hat p_j$ in the target view is given by:
\begin{equation}
    \hat p_{j} = p_i + M_{flow}^{i\rightarrow j}(p_i).
\end{equation}
The true matched point $p_j$ should ideally lie on the epipolar line $l_{p_i}$ in the target view. However, due to estimation errors, the predicted point $\hat p_j$ from the flow estimator often deviates from this condition. To mitigate this, we relax $\hat p_j$ by selecting the perpendicular foot $\bar p_j$ on the epipolar line $l_{p_i}$, which is the closet point to $\hat p_j$. The computation is as follows: 
\begin{equation}
    \bar p_j = (\frac{b^2 \hat x_j - ab \hat y_j - ac}{a^2 + b^2}, \frac{a^2 \hat y_j - ab \hat x_j - bc}{a^2 + b^2}),
\end{equation}
where ($\hat x_j, \hat y_j$) are the coordinates of point $\hat p_j$ in the camera coordinate system, $\{a, b, c\}$ are the parameters of the epipolar line from Eq.~(\ref{eq:epipolar_line}). With $\bar p_j$, we can explicitly compute the depth value of $p_i$ using epipolar geometry:
\begin{equation}
    \begin{aligned}
        D^{i\rightarrow j}(p_i,\bar p_j) = \frac{|H \times -(R_i R_j^{-1} T_j - T_i)|}{|(K_i^{-1} (x_i, y_i, 1)^{\top}) \times H|}, \\
        \textrm{where}~~H = (R_j R_i^{-1})^{-1} K_j^{-1} (\bar x_j, \bar y_j, 1)^{\top}.
        \label{eq:compute_bar_dpt}
    \end{aligned}
\end{equation}
Here, $\times$ denotes the cross product, and $R_i,T_i,R_j,T_j$ are the extrinsic parameters, while $K_i$ and $K_j$ are the intrinsic parameters of the source and target cameras.\\
\textbf{Flow-Resilient Depth Blending.}
\label{sec:depth_blending}
Given a limited number $N$ of input images $\{I_i|i=1,2,\cdots,N\}$, for a pixel $p_i$ in a source view image $I_i$, we can identify the corresponding $N-1$ matched points $\{\hat p_j|j=1,2,\cdots,i-1,i+1,\cdots,N\}$ in the other $N-1$ views and compute the corresponding $N-1$ depth values using Eq.~(\ref{eq:compute_bar_dpt}) with the point $\bar p_j$, which is a relaxation of $\hat{p}_j$. Although $\bar p_j$ lies on the epipolar line, there can still be a distance error between it and the latent true matched point $p_j$. Since this error varies in magnitude across different views, it is necessary to effectively blend these depth values to produce the most accurate depth map.

\begin{figure}
	\centering
	\includegraphics[width=.47\textwidth]{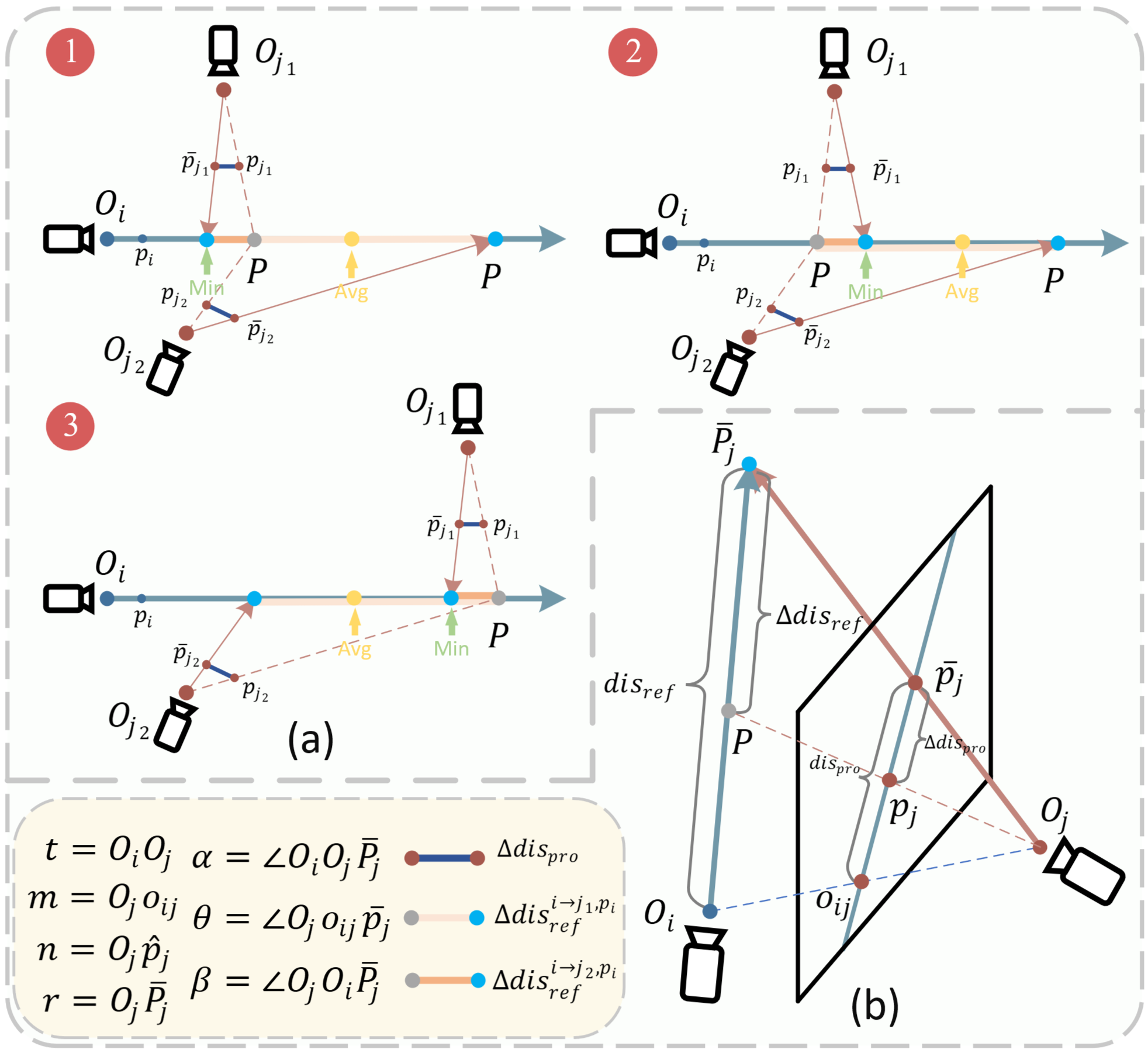}
	\caption{We summarize potential depth blending error scenarios and compare the strategies of selecting the minimum depth versus the average depth. In (a), (1) illustrates the case where two different depth values (blue points) calculated from two views are on either side of the true depth value (gray point), and (2) and (3) illustrate the cases where both depth values lie on the same side of the true depth value, with one on the right and the other on the left, respectively. In (b), we illustrate the geometric relations and notations used to compute $dis_{ref}^\prime(dis_{pro})$.}\vspace{-4mm}
	\label{fig:grad_figure}
\end{figure}

A straightforward approach is to average the $N-1$ depth values. However, as shown in Fig.~\ref{fig:grad_figure} (a), this method is sensitive to errors and often leads to greater deviations from the true 3D corresponding point $P$ compared to selecting the depth value closest to $P$. Nevertheless, since the exact position of $P$ is unknown, reliably identifying the depth value that best matches the true 3D location remains challenging.

We propose a flow-resilient depth blending strategy to address this issue. First, we define a reference distance $dis_{ref}$, which is the distance from the source camera $O_i$ to the 3D point $\bar P_j$, linearly related to depth. We also define a projection distance $dis_{pro}$, corresponding to the length of $dis_{ref}$'s projection onto the target view. As shown in Fig.~\ref{fig:grad_figure} (b), $dis_{pro}$ represents the distance from the epipole $o_{ij}$ to the projected point $\bar p_j$. Since the distance between $\bar p_j$ and the latent true matched point $p_j$ (denoted $\Delta dis_{pro}$) varies across views, any change in the position of $\bar p_j$ will affect depth, thereby altering $dis_{ref}$ and resulting in a distance $\Delta dis_{ref}$. Therefore, we aim to assess the reliability of the depth value computed for a point $p_i$ in each view by examining the rate of change of depth as $\bar p_j$'s position shifts. A smaller rate indicates that the perturbation of $\bar p_j$'s position has minimal impact on depth, making the computed depth for $p_i$ more reliable.

We define $dis_{ref}(dis_{pro})$ as a function of $dis_{pro}$ and compute its gradient, $dis_{ref}^\prime(dis_{pro})$, to quantify this rate of change. Using geometric relations, we derive the following expression for the gradient:
\begin{equation}
	dis_{ref}^\prime(dis_{pro}) = \frac{t\sin{\beta}\sin^2{(\alpha + \theta)}}{m\sin{\theta}\sin^2{(\alpha + \beta)}}.
	\label{eq:gradient}
\end{equation}
The definitions of $\{t,m,\alpha,\theta,\beta\}$ are explained in Fig.~\ref{fig:grad_figure} (b), with the derivation of Eq.~(\ref{eq:gradient}) provided in the supplementary.

According to Eq.~(\ref{eq:gradient}), we apply the Fundamental Theorem of Calculus to convert $\Delta dis_{pro}$ into $\Delta dis_{ref}$, which is expressed as:
\begin{equation}
\begin{aligned}
	\Delta dis_{ref}& = \int_{dis_{pro}}^{dis_{pro}+\Delta dis_{pro}}dis_{ref}^\prime(dis_{pro}) ~\mathrm{d}dis_{pro} \\
	&\approx dis_{ref}^\prime(dis_{pro}) \Delta dis_{pro},
\end{aligned}
\end{equation}
where $\Delta dis_{pro}$ can be either positive or negative. For a point $p_i$ in the source view, we aim to initialize its corresponding 3D point with the depth value that minimizes $\Delta dis_{ref}$ among the other $N-1$ views. Since our goal is to sort $\Delta dis_{ref}$ rather than compute its exact value, we make a reasonable assumption that valid predicted points, when using the same optical flow predictor, have similar absolute values of $\Delta dis_{pro}$. This allows us to ignore the influence of $\Delta dis_{pro}$ and directly compare the gradient values $dis_{ref}^\prime(dis_{pro})$ to evaluate the relationship of $\Delta dis_{ref}$. Finally, the blended depth value $D_i(p_i)$ for the point $p_i$ is determined as follows:
\begin{equation}
	D_i(p_i) = D^{i\rightarrow k}(p_i), k = \arg\min_{j} dis_{ref}^\prime(dis_{pro}^{i\rightarrow j,p_i}).
	\label{11}
\end{equation}

\begin{table*}[t!]
    \centering
        \resizebox{1\linewidth}{!}{ 
            \begin{tabular}{@{}l|cccc|cccc|cccc}
                \toprule
                \multirow{2}{*}{Method}     & \multicolumn{4}{c|}{LLFF (3 Views)}  & \multicolumn{4}{c|}{DTU (3 Views)} & \multicolumn{4}{c}{MipNeRF-360 (24 Views)}    \\
                \cmidrule(lr){2-5}
                 \cmidrule(lr){6-9}
                  \cmidrule(lr){10-13}
                &   PSNR$\uparrow$ & SSIM$\uparrow$ & LPIPS$\downarrow$ & AVGE$\downarrow$ & PSNR$\uparrow$ & SSIM$\uparrow$ & LPIPS$\downarrow$ & AVGE$\downarrow$ & PSNR$\uparrow$ & SSIM$\uparrow$ & LPIPS$\downarrow$ & AVGE$\downarrow$ \\
                \midrule
                Mip-NeRF~\cite{barron2022mip}                      & 14.62 & 0.351 & 0.495 & 0.240 & 8.68  & 0.571 & 0.353 & 0.315 & 21.23 & 0.613 & 0.351 & 0.118\\
                DietNeRF~\cite{jain2021putting}                 & 14.94 & 0.370 & 0.496 & 0.233 & 11.85 & 0.633 & 0.314 & 0.232 & 20.21 & 0.557 & 0.387 &0.135 \\
                RegNeRF~\cite{niemeyer2022regnerf}                  & 19.08 & 0.587 & 0.336 & 0.139 & 18.89 & 0.745 & 0.190 & 0.107 & 22.19 & 0.643 & 0.335 & 0.107\\
                FreeNeRF~\cite{yang2023freenerf}              & 19.63 & 0.612 & 0.308 & 0.128 & \cellcolor{top2}19.92 & 0.787 & 0.182 & 0.095 & 22.78 & 0.689 & 0.323 & 0.098\\
                SparseNeRF~\cite{wang2023sparsenerf}              & 19.86 & 0.624 & 0.328 & 0.128 & \cellcolor{top3}19.55 & 0.769 & 0.201 & 0.102 & 22.85 & 0.693 & 0.315 & 0.097\\ 
                DS-NeRF~\cite{deng2022depth}                 & 19.34 & 0.596 & 0.295 & 0.130 & 16.29 & 0.559 & 0.451 & 0.192 & 14.58 & 0.311 & 0.692 & 0.271\\              
                ViP-NeRF~\cite{somraj2023vip}                & 18.50 & 0.528 & 0.321 & 0.146 & 10.20 & 0.301 & 0.363 & 0.307 & 14.78 & 0.300 & 0.774 & 0.278 \\ 
                SimpleNeRF~\cite{somraj2024simple}                 & 18.82 & 0.587 & 0.274 & 0.132 & 11.94 & 0.387 & 0.286 & 0.243 &10.82 & 0.142 & 0.880 &  0.407 \\ 
                \midrule
                3DGS~\cite{kerbl20233d}                     & 18.54 & 0.588 & 0.272 & 0.135 & 17.65 & 0.816 & \cellcolor{top3}0.146 & 0.102 & 21.71 & 0.672 & 0.248 &0.099 \\
                DNGaussian~\cite{li2024dngaussian}                 & 19.12 & 0.591 & 0.294 & 0.132 & 18.91 & 0.790 & 0.176 & \cellcolor{top3}0.101 & 18.06 & 0.423 & 0.584 &0.191\\
                FSGS~\cite{zhu2023fsgs}                     & \cellcolor{top3}20.43 & \cellcolor{top3}0.682 & \cellcolor{top3}0.248 & \cellcolor{top3}0.108 & 17.14 & \cellcolor{top3}0.818 & 0.162 & 0.110 & \cellcolor{top3}23.40 & \cellcolor{top2}0.733 & \cellcolor{top3}0.238 &\cellcolor{top3}0.083\\             
                CoR-GS~\cite{zhang2024cor}                     & \cellcolor{top2}20.45 & \cellcolor{top2}0.712 & \cellcolor{top2}0.196 & \cellcolor{top2}0.098 & 19.21 & \cellcolor{top2}0.853 & \cellcolor{top2}0.119 & \cellcolor{top2}0.082 & \cellcolor{top2}23.55 & \cellcolor{top3}0.727 & \cellcolor{top2}0.226 & \cellcolor{top2}0.080 \\              
                \textbf{Ours}            & \cellcolor{top1}21.07 & \cellcolor{top1}0.738 & \cellcolor{top1}0.177 & \cellcolor{top1}0.089 &  \cellcolor{top1}20.21 & \cellcolor{top1}0.869 & \cellcolor{top1}0.102  & \cellcolor{top1}0.071 & \cellcolor{top1}23.86 & \cellcolor{top1}0.753 & \cellcolor{top1}0.206 & \cellcolor{top1}0.075\\
                \bottomrule
            \end{tabular}
            }
        \caption{Quantitative evaluations on the LLFF, DTU and Mip-NeRF360 datasets. For DNGaussian and FSGS training on DTU, we initialize with random point clouds when COLMAP fails due to sparse input images. The \colorbox[RGB]{255,179,179}{best}, \colorbox[RGB]{255,217,179}{second best}, and \colorbox[RGB]{255,255,179}{third best} results are color-coded in each cell.}
        \label{tab:llff_dtu}\vspace{-6mm}    
    \end{table*}

\noindent\textbf{Flow-Filtered Depth Pruning.}
Since optical flow estimation may fail due to color similarity or pixel deviation from the target view, the assumption regarding similar projection distance errors may not always hold. In such cases, these points with unreliable flow need to be discarded. To achieve this, we compute the distance $g^{i\rightarrow j}(p_i)$ from the predicted matched point $\hat p_j$ to the epipolar line $l_{p_i}$: 

\begin{equation}
    g^{i\rightarrow j}(p_i) = \frac{|a\hat x_j + b\hat y_j + c|}{\sqrt{a^2 + b^2}}.
    \label{eq:distance}
\end{equation}
A larger distance indicates lower confidence in the optical flow estimation for that point. Therefore, we set a threshold $\epsilon_g$ to filter out points with large $g^{i\rightarrow j}(\cdot)$. This not only improves the quality of the point cloud by reducing the influence of erroneous points but also reinforces the reliability of our assumption on $\Delta dis_{pro}$. 

Next, we assign the color of each pixel from the input views as the initial color of the corresponding Gaussian, and use the calculated depth to project the pixel into 3D space, establishing the initial position $\mu$ for each Gaussian. Combining Eq.~(\ref{11}) and (\ref{eq:distance}), we compute $\mu$ as follows:
\begin{equation}
    \mu = \mathcal{P}(p_i, D_i(p_i)), \text{ where } g^{i\rightarrow k}(p) < \epsilon_d,
\end{equation}
where $\mathcal{P}(\cdot)$ is the projection function that maps 2D points into 3D space using the corresponding depth value.

\noindent\textbf{Objectives.}
During training, we use Eq.~(\ref{eq:render}) to generate the image $I_{render}$, and then compute the difference between $I_{render}$ and the ground truth image $I_{gt}$ using a loss function composed of L1 and D-SSIM:
\begin{equation}
    \mathcal{L} = (1-\lambda_{c}) \mathcal{L}_{1} + \lambda_{c} \mathcal{L}_{D-SSIM}.
    \label{eq:rgb_loss}
\end{equation}
Thanks to the dense and accurate initial point cloud with epipolar depth priors, our approach does not require additional depth regularization in the objective function.

\vspace{-2mm}\section{Experiments}
\subsection{Experimental Settings}
\noindent\textbf{Datasets.} 
We evaluate our method on the LLFF~\cite{mildenhall2019local}, Blender~\cite{mildenhall2021nerf}, DTU~\cite{jensen2014large}, and Mip-NeRF360~\cite{barron2022mip} datasets, using the same splits and resolution as previous works~\cite{niemeyer2022regnerf, wang2023sparsenerf, yang2023freenerf, zhu2023fsgs, li2024dngaussian}. Among these, LLFF is a real-world dataset focused on forward-facing scenes, Mip-NeRF360 is a large-scale dataset for 360-degree scenes, and DTU and Blender focus on object-centric datasets. For DTU, we apply masks to remove background influence, following prior methods~\cite{niemeyer2022regnerf}. Camera parameters are assumed to be known, consistent with other sparse view synthesis approaches~\cite{zhu2023fsgs, li2024dngaussian, zhang2024cor}.

\noindent\textbf{Competitors.}
We compare our method against state-of-the-art approaches, including eight NeRF-based methods, such as SparseNeRF~\cite{wang2023sparsenerf}, DS-NeRF~\cite{deng2022depth}, ViP-NeRF~\cite{somraj2023vip}, and SimpleNeRF~\cite{somraj2024simple}, as well as recent 3DGS-based methods like DNGaussian~\cite{li2024dngaussian}, FSGS~\cite{zhu2023fsgs}, CoR-GS~\cite{zhang2024cor}, and the standard 3DGS~\cite{kerbl20233d}. Unless noted otherwise, all reported results are either taken from the original papers or obtained by training using the official code and default settings.

\noindent\textbf{Metrics.}
Consistent with previous methods~\cite{niemeyer2022regnerf, yang2023freenerf, wang2023sparsenerf, li2024dngaussian, zhu2023fsgs, zhang2024cor}, we use PSNR, SSIM, and LPIPS as evaluation metrics. Additionally, we compute the AVGE metric~\cite{niemeyer2022regnerf}, which is the geometric mean of PSNR, SSIM, and LPIPS.

\begin{figure*}[!t]
	\setlength{\abovecaptionskip}{12pt}
	\centering
	\includegraphics[width=1\linewidth]{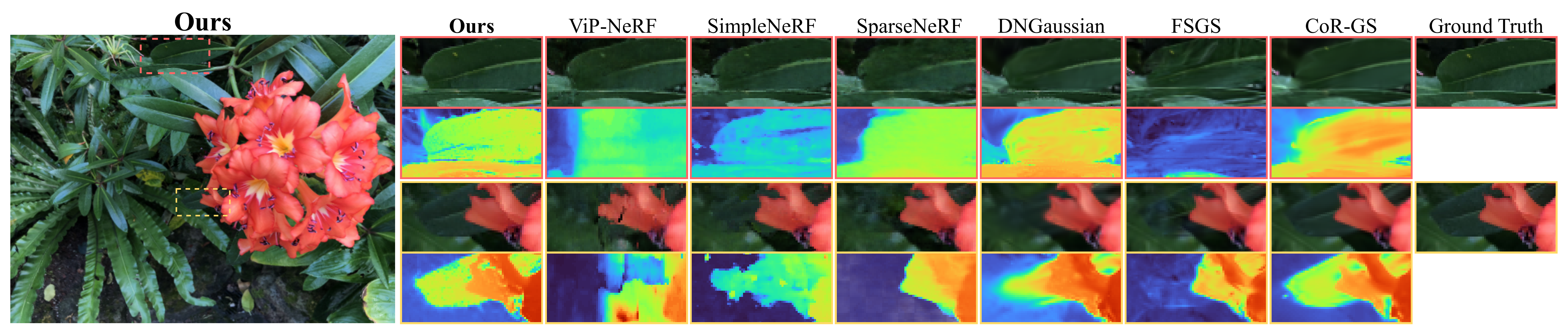}
	\vspace{-6mm}\caption{
		Visual comparisons on LLFF (3 views). Our method produces richer details and more accurate depth than all competitors. 
	}\vspace{-2mm}
	\label{fig:llff_result}
\end{figure*}

\subsection{Comparison with State-of-the-Art Methods}
\noindent\textbf{LLFF}. 
We summarize the quantitative results in Tab.~\ref{tab:llff_dtu}. Our method outperforms all others in PSNR, SSIM, LPIPS, and AVGE on the LLFF dataset, achieving significant improvements of 0.62, 0.026, 0.019, and 0.009, respectively, over the second-best method.

Fig.~\ref{fig:llff_result} shows the visual comparisons, where our approach excels in both geometry and texture reconstruction. NeRF-based methods like ViP-NeRF struggle with geometry due to the lack of depth priors, as seen in the poor shape reconstruction. SimpleNeRF, using filtered depth priors, achieves better geometry but suffers from sparse depth, leading to incomplete reconstructions, especially at object edges. While DNGaussian and FSGS capture more texture detail, they are hindered by geometric inaccuracies from unreliable monocular depth maps, resulting in missing Gaussians and incomplete object reconstructions. Although CoR-GS has relatively complete depth maps, the overall reconstruction remains overly smooth and fails to capture high-frequency texture details. In contrast, our method, leveraging a dense and accurate point cloud with epipolar depth priors, not only reconstructs the full geometry but also captures finer details more effectively than the other methods.

\begin{figure*}[!t]
    \setlength{\abovecaptionskip}{12pt}
    \centering
    \includegraphics[width=1\linewidth]{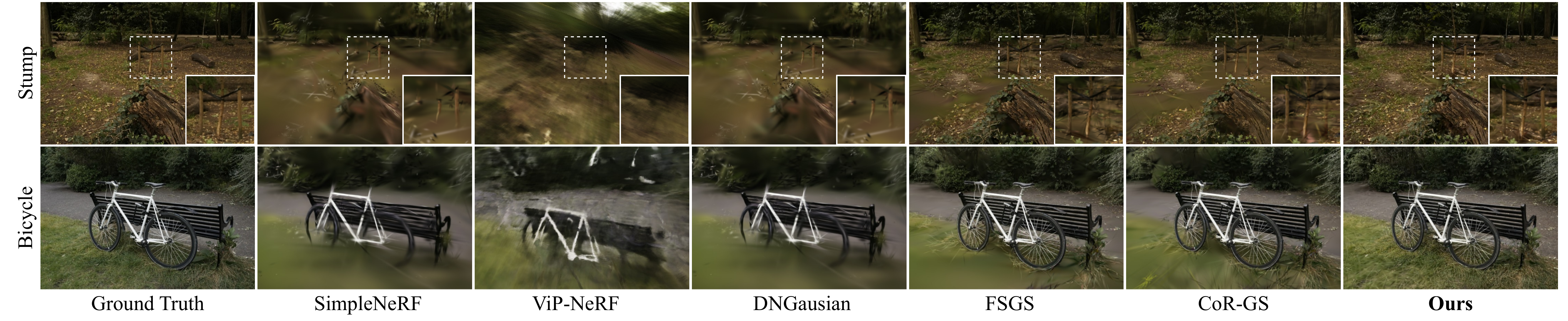}\vspace{-4mm}
    \caption{
        Visual comparisons on MipNeRF-360 (24 views). In large-scale scenes, NexusGS preserve detailed textures while others struggle.
    }\vspace{-4mm}
    \label{fig:mip_result}
\end{figure*}
\noindent\textbf{MipNeRF-360}. 
In the large-scale MipNeRF-360 dataset, obtaining accurate and dense point clouds under sparse-view settings is more challenging. Qualitative results are shown in Fig.~\ref{fig:mip_result}. DNGaussian, being fast and lightweight, struggles with unbounded scenes. NeRF-based methods using depth priors, such as SimpleNeRF and ViP-NeRF, suffer from unreliable and sparse depth. Other 3DGS methods like FSGS and CoR-GS, while achieving good scores, still exhibit over-smoothing in large-scale scenes due to insufficient point cloud density. This demonstrates that initializing with sparse point clouds under limited supervision is inadequate for densification. In contrast, directly generating dense point clouds enables more complete scene reconstruction and finer details. As shown in Tab.~\ref{tab:llff_dtu}, our method achieves state-of-the-art results across all metrics.
\begin{figure}
	\centering
	\includegraphics[width=.47\textwidth]{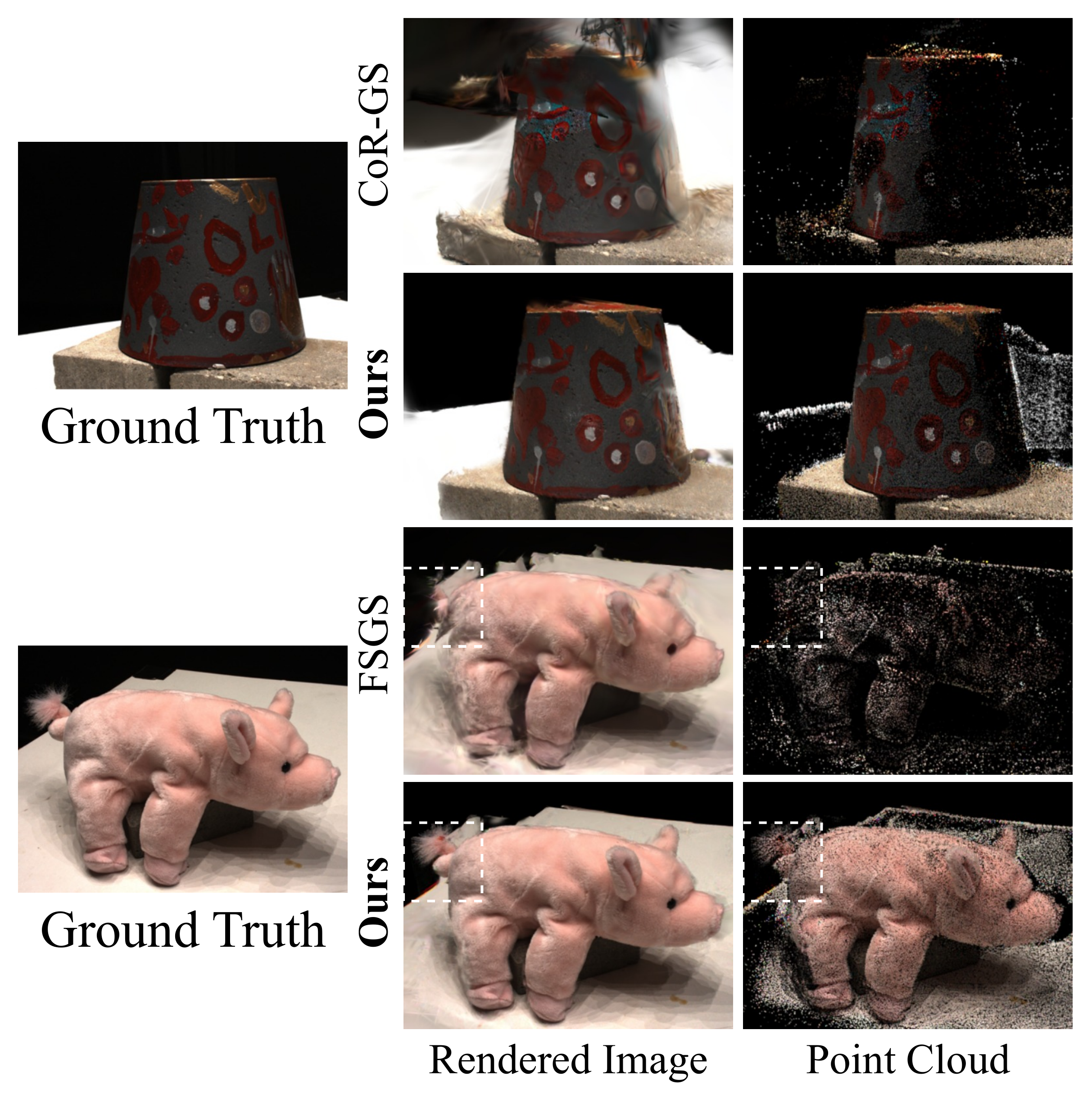}\vspace{-2mm}
	\vspace{-2mm}\caption{Visual comparisons on the DTU dataset (3 views). Our method produces a more comprehensive point cloud than others, resulting in higher-quality renderings.}\vspace{-6mm}
	\label{fig:dtu_result}
\end{figure}

\begin{table}[t]
    \centering
    \resizebox{0.95\linewidth}{!}{
    \begin{tabular}{l|cccc}
    \toprule
    \multirow{2}{*}{Method} & \multicolumn{4}{c}{Blender (8 Views)}                           \\
     \cmidrule(lr){2-5}
        & PSNR $\uparrow$        & SSIM $\uparrow$        & LPIPS $\downarrow$ & AVGE $\downarrow$   \\
    \midrule
    NeRF~\cite{mildenhall2021nerf}                                     & 14.93          & 0.687          & 0.318   & 0.179              \\
    Simplified NeRF~\cite{jain2021putting}                     & 20.09          & 0.822          & 0.179     & 0.090               \\
    DietNeRF~\cite{jain2021putting}                             & 23.15          & 0.866          & 0.109    & 0.058               \\
    DietNeRF + ft~\cite{jain2021putting}                        & 23.59          & 0.874          & 0.097    & 0.053              \\
    FreeNeRF~\cite{yang2023freenerf}                            & \cellcolor{top2}24.26          & 0.883          & 0.098    & \cellcolor{top3}0.050              \\
    SparseNeRF~\cite{wang2023sparsenerf}                        & 22.41          & 0.861          & 0.119    & 0.063             \\
    \midrule
    3DGS~\cite{kerbl20233d}                                     & 22.23          & 0.858          & 0.114    & 0.064               \\
    FSGS~\cite{zhu2023fsgs}                                                        & 22.70          & 0.822          & 0.137    &  0.068              \\
	CoR-GS~\cite{zhang2024cor}                          & 23.98          & \cellcolor{top2}0.891          & \cellcolor{top3}0.094    &  \cellcolor{top3}0.050             \\
    DNGaussian~\cite{li2024dngaussian}                          & \cellcolor{top3}24.07          & \cellcolor{top3}0.885          & \cellcolor{top2}0.090    &  \cellcolor{top2}0.049              \\
    \textbf{Ours}                           & \cellcolor{top1}24.37& \cellcolor{top1}0.893 & \cellcolor{top1}0.087 & \cellcolor{top1}0.047\\
    \bottomrule
    \end{tabular}
    } 
   \vspace{-2mm} \caption{
        Quantitative evaluations on the Blender dataset.
    }\vspace{-7mm}
    \label{table:blender}
\end{table} 

\noindent\textbf{DTU.} 
In this dataset, our visual comparisons, shown in Fig.~\ref{fig:dtu_result}, focus on comparing NexusGS with other 3DGS-based methods. Our reconstructed point cloud is denser and more accurate than those of CoR-GS and FSGS, which use the original 3DGS densification process and produce sparse or incomplete clouds. NexusGS also delivers superior rendering quality in terms of detail and texture. This highlights the effectiveness of initializing with a dense point cloud, rather than densifying from sparse views. Quantitative results can be found in Tab.~\ref{tab:llff_dtu}, with additional visual comparisons in the supplementary materials.

\noindent\textbf{Blender.}
We achieve state-of-the-art quantitative results on this dataset (Tab.~\ref{table:blender}). Despite the synthetic nature and lower demands for fine details, our method outperforms previous NeRFs, thanks to effective depth prior integration. While DNGaussian yields good results, it lacks the density of our point cloud. Additionally, CoR-GS suffers from degraded performance due to the absence of depth priors, whereas FSGS struggles to integrate them effectively, relying on a relaxed Pearson coefficient loss. Qualitative results are provided in the supplementary materials.

\subsection{Ablation Study}
\begin{table}[!t]
	\centering
	\resizebox{1.0\linewidth}{!}{	
		\begin{tabular}{@{}c|cc|cccc}
			\toprule		
			& \multicolumn{2}{c|}{Setting} & PSNR $\uparrow$ & SSIM $\uparrow$  & LPIPS $\downarrow$ & AVGE $\downarrow$  \\
			\midrule
			3DGS  &  \multicolumn{2}{c|}{/} & 18.54 & 0.588 & 0.272 & 13.469\\ \midrule
			\multirow{8}{*}{\textbf{Ours}}       & \multicolumn{2}{l|}{\textbf{A}. Average} & 20.39 & 0.722 & 0.186 & 9.642 \\
			& \multicolumn{2}{l|}{\textbf{B}. Nearest} & 20.62 & 0.735 & 0.178 & 9.261 \\
			& \multicolumn{2}{l|}{\textbf{C}. Weighted Average} & 20.59 & 0.732 & 0.180 & 9.335 \\
			& \multicolumn{2}{l|}{\textbf{D}. FRDB} & 20.90 & 0.734 & 0.180 & 9.104\\
			& \multicolumn{2}{l|}{\textbf{E}. Average + FFDP} & 20.51 & 0.725 & 0.184 & 9.502 \\
			& \multicolumn{2}{l|}{\textbf{F}. Nearest + FFDP} & 20.79 & 0.734 & 0.178 & 9.147 \\
			& \multicolumn{2}{l|}{\textbf{G}. Weighted + FFDP} & 20.78 & 0.735 & 0.178 &  9.148 \\	
			& \multicolumn{2}{l|}{\textbf{H}. FRDB + FFDP} & \textbf{21.07} & \textbf{0.738} & \textbf{0.177} & \textbf{8.913} \\
			\bottomrule
		\end{tabular}
	}
	\vspace{-3mm}\caption{Ablation study on LLFF with 3 training views. To clarify the comparison, the AVGE values are multiplied by 1e2.}\vspace{-4mm}
	\label{tab:ablation}
\end{table}

\begin{figure}
	\centering
	\includegraphics[width=.49\textwidth]{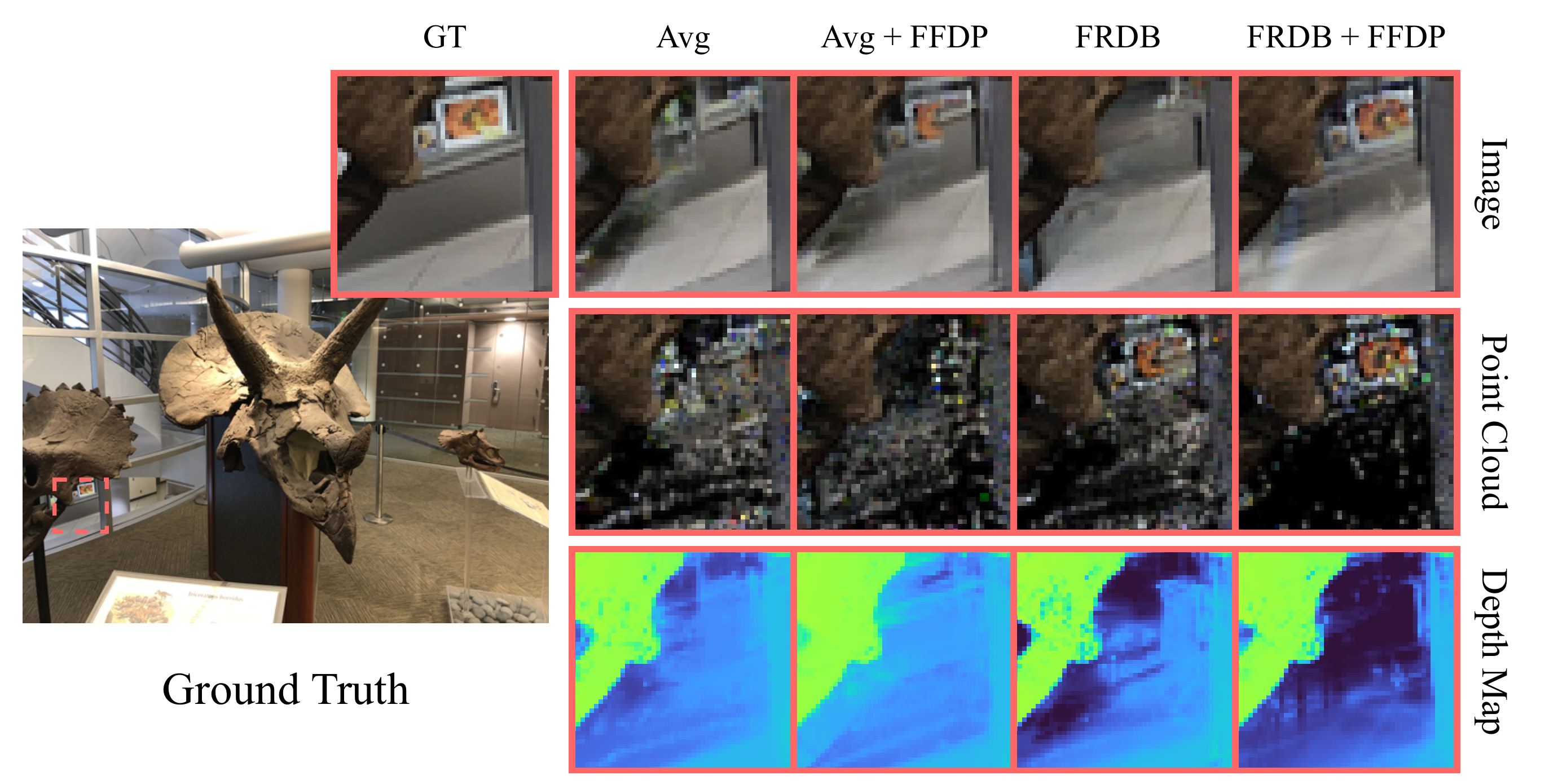}\vspace{-4mm}
	\caption{Visualization of ablation study results using 3-views.}\label{fig:ablation}\vspace{-6mm}
\end{figure}

We conduct ablation experiments on LLFF dataset with 3 views to assess the impact of Flow-Resilient Depth Blending (FRDB) and Flow-Filtered Depth Pruning (FFDP). We compare four depth blending strategies: Strategy \textbf{A} uses a simple average, \textbf{B} uses the depth from the nearest target view, and \textbf{C} employs a weighted average based on the inverse epipolar distance. FRDB (\textbf{D}) outperforms all others, providing superior reconstruction quality. Tab.~\ref{tab:ablation} shows that applying FFDP further improves all strategies.

In Fig.~\ref{fig:ablation}, we compare two representative strategies, \textbf{A} (Average) and \textbf{D} (FRDB). The mean strategy produces reasonable reconstructions but introduces errors in finer details, as seen in the point cloud, where erroneous points obscure the correct ones. During sparse-view training, these errors cannot be filtered out, leading to incorrect renderings. In contrast, FRDB with FFDP yields more accurate depth and a cleaner point cloud, significantly enhancing reconstruction. A comparison of depth maps highlights our method's superior depth accuracy.

\section{Generalizability Analysis}
To evaluate the generalizability of the depth maps and initial point clouds generated by NexusGS, we integrate them into existing methods and assess their performance on the LLFF dataset, with results summarized in Tab.~\ref{tab:transfer}.

First, incorporating our depth maps as depth priors into SparseNeRF~\cite{wang2023sparsenerf} yields significant improvements, showing their effectiveness in enhancing NeRF-based methods.

For 3DGS-based methods, both our depth maps and point clouds are effective. Applying either to DNGaussian~\cite{li2024dngaussian} lead to notable improvements, similar to the gains seen with SparseNeRF. Compared to using fused stereo point clouds (DNGaussian*), our point clouds improve SSIM and LPIPS more than PSNR, likely due to DNGaussian’s reliance on inaccurate monocular depth maps for regularization. Nonetheless, our dense point clouds reduce over-smoothing, improving perceptual scores. Using only our depth maps improves PSNR, but DNGaussian's point clouds still lack sufficient density, which may result in some over-smoothing. When both depth maps and point clouds are used together, or when point clouds are used without depth loss, the depth information is better leveraged, resulting in improved SSIM and LPIPS scores. These configurations yield similar results, as the point clouds are derived directly from the depth maps.
\begin{table}[t]	
		\resizebox{0.47\textwidth}{!}{
			\begin{tabular}{l|ccc}
				\toprule
				Method (3 views) & PSNR $\uparrow$        & SSIM $\uparrow$        & LPIPS $\downarrow$ \\ \midrule
				SparseNeRF & 19.86    & 0.624    & 0.328  \\
				SparseNeRF + Our Depth &  \textbf{20.23}   &  \textbf{0.647}   &   \textbf{0.313}   \\
				\midrule
				DNGaussian & 19.12    & 0.591    & 0.294       \\
				DNGaussian* & 19.96    & 0.684    & 0.232       \\
				DNGaussian + Our Point Cloud (PC) & 19.97    & 0.713    & 0.210     \\
				DNGaussian + Our Depth & 20.09    & 0.694    & 0.220     \\
				DNGaussian + Our PC\&Depth & \textbf{20.11}    & \textbf{0.723}    & \textbf{0.200}     \\
				DNGaussian w/o Depth + Our PC & 20.10    & 0.719    & 0.202     \\ \midrule
				FSGS & 20.43    & 0.682    & 0.248       \\
				FSGS + Our PC & \textbf{20.85}    & \textbf{0.746}    & \textbf{0.166}       \\ \midrule
				CoR-GS & 20.45    & 0.712    & 0.196       \\
				CoR-GS + Our PC & \textbf{21.00}    & \textbf{0.744}    & \textbf{0.176}       \\ \bottomrule			
			\end{tabular}
		}		
		\vspace{-2mm}\caption{Versality of the depth map and initialized point cloud from our method. The best result for each method is highlighted in bold.}\vspace{-4mm}
		\label{tab:transfer}
	\end{table}
	
For FSGS~\cite{zhu2023fsgs}, which requires depth regularization on pseudo-views, we cannot apply our depth maps due to the difficulty in generating pseudo-view depths. However, using only our point clouds still yields significant improvements, further demonstrating the generalizability of our approach.

CoR-GS~\cite{zhang2024cor} optimizes two 3DGS radiance fields with co-regularization, constraining their rendering results with pseudo-views. Replacing the initial point cloud of one radiance field with ours leads to substantial improvement. The priors from our point cloud transfer effectively, combining the benefits of both the fused stereo point cloud and our dense point cloud. These results highlight the broad applicability and effectiveness of our method.

\section{Conclusion}
This paper presents NexusGS, a method that enhances sparse view synthesis in 3D Gaussian Splatting by embedding epipolar depth priors. Our approach uses optical flow and camera poses to compute accurate depth, yielding a reliable initial dense point cloud that improves rendering fidelity. Key components include Epipolar Depth Nexus for precise depth calculation, Flow-Resilient Depth Blending to reduce flow-induced errors, and Flow-Filtered Depth Pruning to eliminate inconsistent points. Experiments show that NexusGS outperforms state-of-the-art methods in depth accuracy and rendering quality, demonstrating the effectiveness of integrating epipolar depth information in sparse-view 3DGS.

%\noindent\textbf{Limitation.} Despite NexusGS's remarkable performance and generalizability, it, like most sparse-view synthesis methods, relies on known camera poses to implement epipolar constraints. While recent pose-free methods, such as COGS~\cite{jiang2024construct}, bypass this requirement, the balance between pose-free flexibility and rendering accuracy shows room for improvement and offers a promising avenue for future work.

\small{\noindent\textbf{Acknowledgements.} This work is supported by the National Natural Science Foundation of China (No. 62102381, 41927805); Shandong Natural Science Foundation (No. ZR2021QF035); the National Key R\&D Program of China (No. 2022ZD0117201); the Guangdong Natural Science Funds for Distinguished Young Scholar (No. 2023B1515020097); the AI Singapore Programme under the National Research Foundation Singapore (Grant AISG3-GV-2023-011); and the Lee Kong Chian Fellowships.}

\maketitlesupplementary

\section{Derivations}
\subsection{Epipolar Depth Nexus}
Once the coordinates of the perpendicular foot, denoted as $\bar p_j$, are determined in the target view's camera coordinate system (as shown in Eq.~(7) in the main paper), the depth of point $p_i$, denoted as $D^{i\rightarrow j}(p_i,\bar p_j)$, can be computed using Eq.~(8). Below, we provide a detailed derivation of Eq.~(8) as presented in the main paper.

Since both $p_i$ and $\bar p_j$ are expressed in their respective camera coordinate systems, we first convert them into normalized image coordinates to facilitate the depth calculation. These coordinates, situated in their respective 3D domains, are referenced with the source view camera $O_i$ and the target view camera $O_j$ as origins. The transformation is formulated as follows:
\begin{equation}
	\tilde p_i = K_i^{-1} (x_i, y_i, 1)^{\top},~~~~\tilde p_j = K_j^{-1} (\bar x_j, \bar y_j, 1)^{\top},
\end{equation}
where $K_i$ and $K_j$ represent the intrinsic parameters of the source and target cameras, while $(x_i,y_i),(\bar x_j, \bar y_j)$ are the image coordinates of points $p_i$ and $\bar p_j$, respectively.

\begin{figure}
	\centering
	\includegraphics[width=.45\textwidth]{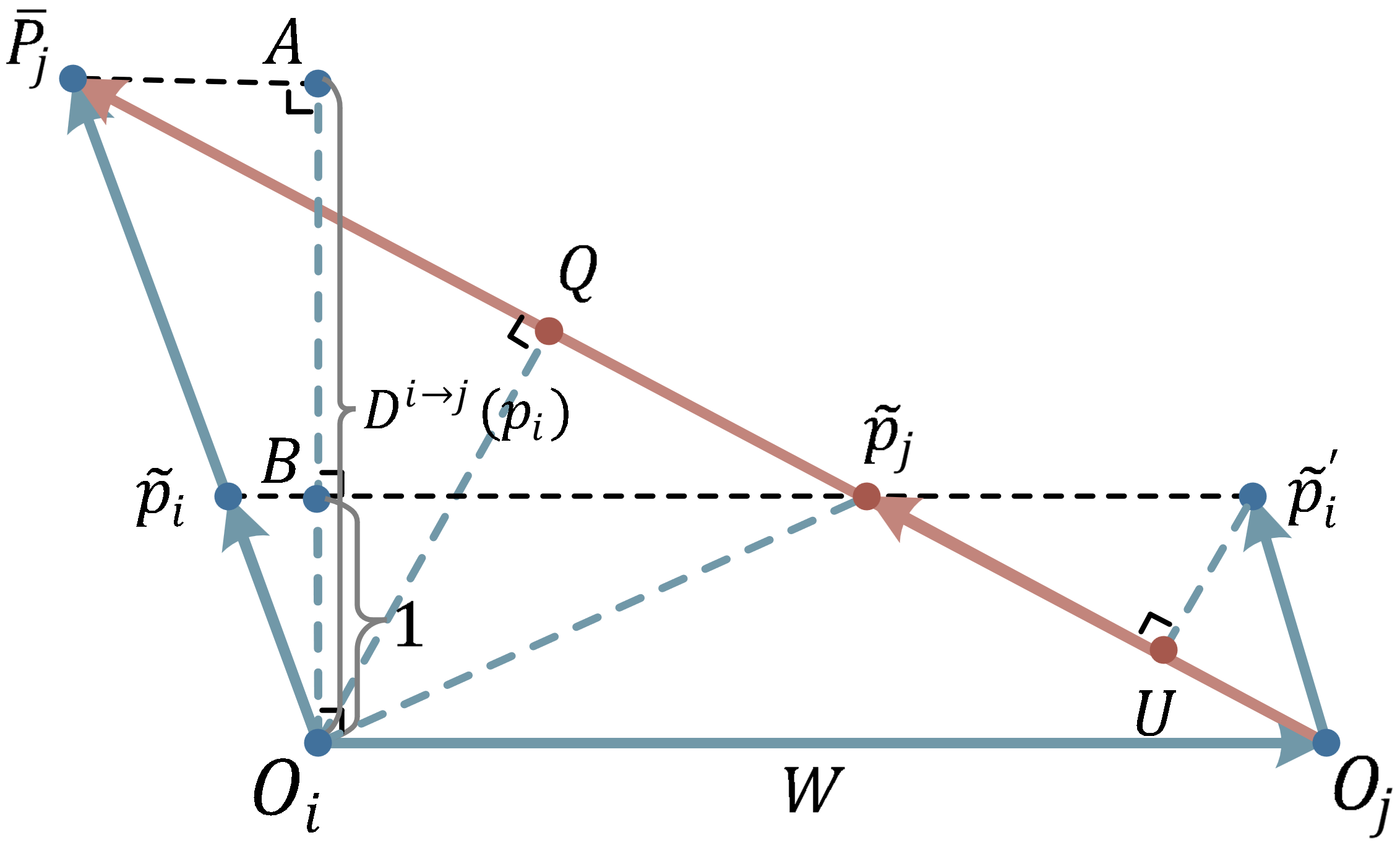}
	\caption{Illustration of the geometry relationships used in Epipolar Depth Nexus step.}
	\label{fig:supp1_figure}
\end{figure}

Next, we refer to Fig.~\ref{fig:supp1_figure} to elucidate the geometric relationships employed in our method. Notably, the triangles $\triangle O_iA \bar P_j$ and $\triangle O_iB\tilde{p}_i$ are similar, allowing us to establish the following relationship:
\begin{equation}
	\frac{|\overrightarrow{O_i \bar P_j}|}{|\overrightarrow{O_i \tilde p_i}|} = \frac{|O_i A|}{|O_i B|} = \frac{D^{i\rightarrow j}(p_i)}{1}.
	\label{equ:2}
\end{equation}

Here, $|\overrightarrow{O_i \bar P_j}|$ and $|\overrightarrow{O_i \tilde p_i}|$ represent the magnitudes of the respective vectors. Similarly, $|O_i A|$ and $|O_i B|$ represent the respective distances from $O_i$ to the planes containing $\bar P_j$ and $\tilde p_i$ along the optical axis.

To further analyze the geometry, we introduce an auxiliary point $\tilde p_i^{'}$ such that $\overrightarrow{O_j\tilde p_i^{'}}$ has the same length and direction as $\overrightarrow{O_i\tilde p_i}$. Projecting $O_i$ perpendicularly onto the line $O_j \bar P_j$ yields point $Q$, and projecting $\tilde p_i^{'}$ onto the same line yields point $U$. The alternate interior angles $\angle O_i \bar P_j Q$ and $\angle \tilde p_i^{'}O_j U$ are equal, leading to similar triangles $\triangle O_i \bar P_j Q$ and $\triangle \tilde p_i^{'}O_j U$. Notice that triangles $\triangle O_i\tilde p_jO_j$ and $\triangle \tilde p_i^{'}O_j\tilde p_j$ share the same base, allowing us to further derive Eq.~(\ref{equ:2}) as follows:
\begin{equation}
	D^{i\rightarrow j}(p_i) = \frac{|\overrightarrow{O_i \bar P_j}|}{|\overrightarrow{O_i \tilde p_i}|} = \frac{|O_i Q|}{|\tilde p_i^{'} U|} = \frac{\textrm{Area}(\triangle O_i\tilde p_jO_j)}{\textrm{Area}(\triangle \tilde p_i^{'}O_j\tilde p_j)}.
\end{equation}
Using the formula for the area of a triangle, the above equation simplifies to:
\begin{equation}
	D^{i\rightarrow j}(p_i) = \frac{|\overrightarrow{O_j \tilde p_j} \times \overrightarrow{O_j O_i}|}{|\overrightarrow{O_j \tilde p_i^{'}} \times \overrightarrow{O_j \tilde p_j}|}.\label{equ:4}
\end{equation}
The 3D coordinates of vectors $\overrightarrow{O_j \tilde p_j}$, $\overrightarrow{O_j O_i}$, and $\overrightarrow{O_j \tilde p_i^{'}}$ can be determined using the extrinsic parameter transformation formulas:
\begin{equation}
	\begin{aligned}
		\overrightarrow{O_j \tilde p_j}
		&= \overrightarrow{O_i \tilde p_j} - \overrightarrow{O_i O_j} \\
		&= (R_j R_i^{-1})^{-1} \tilde p_j + W - W \\
		&= (R_j R_i^{-1})^{-1} K_j^{-1} (\bar x_j, \bar y_j, 1)^{\top},
	\end{aligned}\label{equ:5}
\end{equation}
\begin{equation}
	\overrightarrow{O_j O_i} = -W= -(R_i R_j^{-1} T_j - T_i),\label{equ:6}
\end{equation}
\begin{equation}
	\overrightarrow{O_j \tilde p_i^{'}} = \overrightarrow{O_i \tilde p_i} = \tilde p_i,\label{equ:7}
\end{equation}
where $R_i,T_i$ and $R_j,T_j$ are the extrinsic parameters of the source and target cameras. Finally, we plug Eqs.~(\ref{equ:5}), (\ref{equ:6}) and (\ref{equ:7}) into Eq.~(\ref{equ:4}) to obtain the analytical form of the depth of point $p_i$ as follows:
\begin{equation}
	\begin{aligned}
		D^{i\rightarrow j}(p_i) = \frac{|H \times -(R_i R_j^{-1} T_j - T_i)|}{|(K_i^{-1} (x_i, y_i, 1)^{\top}) \times H|}, \\
		\textrm{where}~~H = (R_j R_i^{-1})^{-1} K_j^{-1} (\bar x_j, \bar y_j, 1)^{\top}.
	\end{aligned}
\end{equation}

\begin{figure}
	\centering
	\includegraphics[width=.45\textwidth]{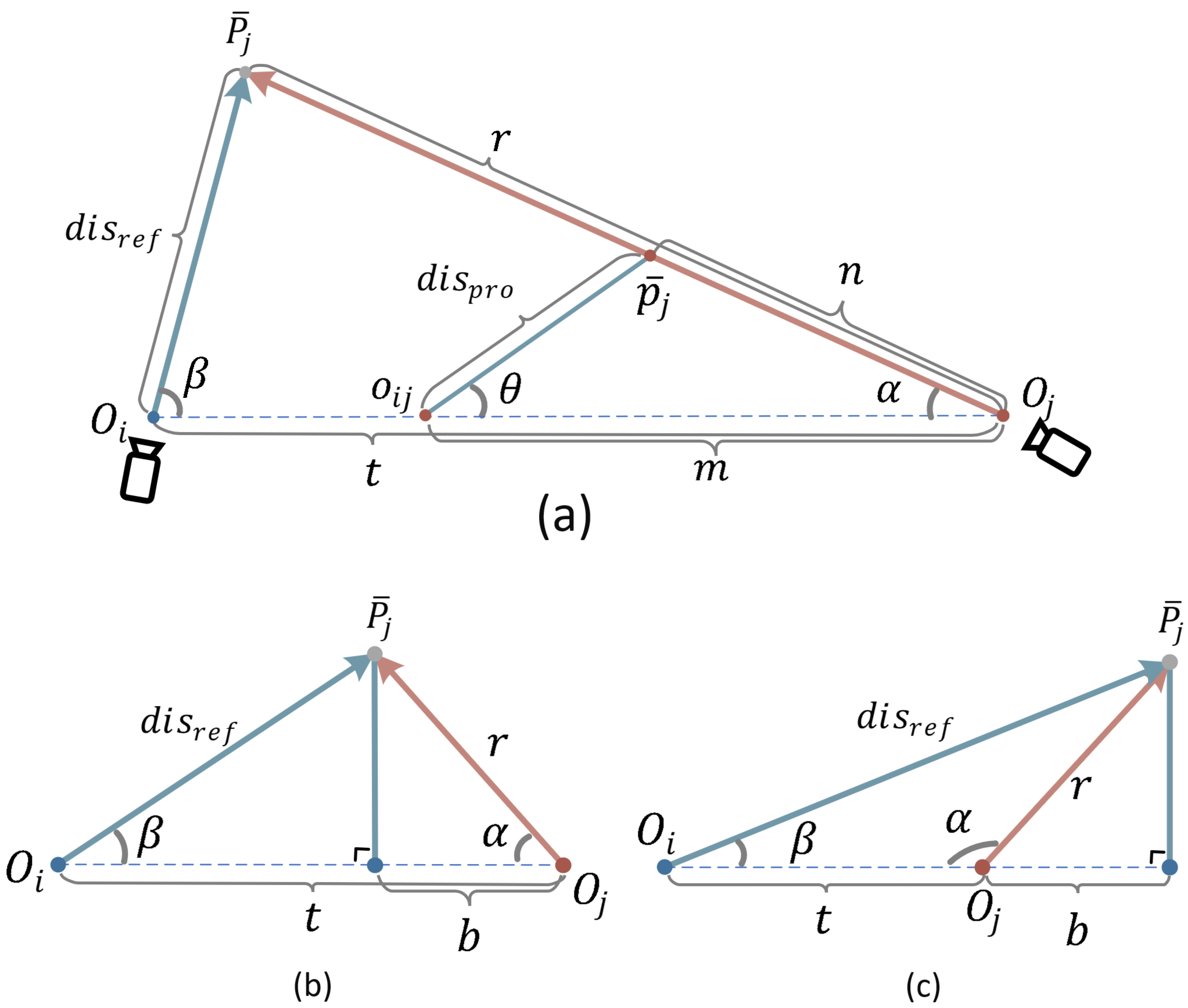}
	\caption{Illustration of the geometric definitions used.}
	\label{fig:supp2_figure}
\end{figure}

\subsection{Flow-Resilient Depth Blending}
Here, we derive Eq.~(9) from the main paper, used in our Flow-Resilient Depth Blending technique. Fig.~\ref{fig:supp2_figure} illustrates the geometric definitions of all relevant symbols. Our objective is to obtain the analytical form of the derivative of $dis_{ref}$ with respect to $dis_{pro}$, denoted as $dis_{ref}^\prime(dis_{pro})=\frac{\mathrm{d} dis_{ref}}{\mathrm{d}dis_{pro}}$. This derivative is decomposed into two factors, $\frac{\mathrm{d}dis_{ref}}{\mathrm{d}\alpha}$ and $\frac{\mathrm{d}\alpha}{\mathrm{d}dis_{pro}}$, connected by the chain rule.

\noindent\textbf{Derivation of $\frac{\mathrm{d}dis_{ref}}{\mathrm{d}\alpha}$.} We begin by determining the expression of $\frac{\mathrm{d}dis_{ref}}{\mathrm{d}\alpha}$. Using the law of cosines and the law of sines, and referring to Fig.~\ref{fig:supp2_figure}~(a) (similar to Fig. 3~(b) in the main paper), we derive the following equations:
\begin{equation}
	\cos{\beta} = \frac{dis_{ref}^2 + t^2 - r^2}{2dis_{ref}t},\label{equ:9}
\end{equation}
\begin{equation}
	r = \frac{dis_{ref} \sin{\beta}}{\sin{\alpha}}.\label{equ:10}
\end{equation}
Substituting Eq.~(\ref{equ:10}) into Eq.~(\ref{equ:9}) and simplifying yields:
\begin{equation}
	\begin{aligned}
		\frac{dis_{ref}^2 \sin^2{\beta}}{\sin^2{\alpha}} &= dis_{ref}^2 + t^2 -2dis_{ref}t\cos{\beta} \\
		&= dis_{ref}^2\sin^2{\beta} + dis_{ref}^2\cos^2{\beta} \\
		&\ \ \ \ + t^2 -2dis_{ref}t\cos{\beta} \\
		&= dis_{ref}^2\sin^2{\beta} + (dis_{ref}\cos{\beta} - t)^2.\label{equ:11}
	\end{aligned}
\end{equation}
Furthermore, we rewrite Eq.~(\ref{equ:11}) in the following form:
\begin{equation}
	\begin{aligned}
		(dis_{ref}\cos{\beta} - t)^2 &= dis_{ref}^2\sin^2{\beta}(\frac{1}{\sin^2{\alpha}}-1) \\
		&=\frac{dis_{ref}^2\sin^2{\beta}\cos^2{\alpha}}{\sin^2{\alpha}}.\label{equ:12}
	\end{aligned}
\end{equation}
Next, we isolate $\frac{\cos^2{\alpha}}{\sin^2{\alpha}}$ from the right-hand side of Eq.~(\ref{equ:12}) to get:
\begin{equation}
	\frac{\cos^2{\alpha}}{\sin^2{\alpha}}=(\frac{\cos{\beta}}{\sin{\beta}}-\frac{t}{dis_{ref}\sin{\beta}})^2.
\end{equation}
To take the square root of both sides of the equation, we need to ascertain the positive or negative nature of each side under the square root. Specifically, the formula after taking the square root can be determined based on geometric relationships. When $\frac{\cos\alpha}{\sin\alpha} > 0$, indicating that $\alpha$ is an acute angle, as illustrated in Fig.~\ref{fig:supp2_figure} (b), the following formulations hold:
\begin{equation}
	\frac{\cos{\beta}}{\sin{\beta}} = \frac{t-b}{dis_{ref}\sin\beta},
\end{equation}
\begin{equation}
	\frac{\cos{\beta}}{\sin{\beta}}-\frac{t}{dis_{ref}\sin{\beta}} = -\frac{b}{dis_{ref}\sin\beta} < 0,~~ \textrm{s.t.}~\frac{\cos\alpha}{\sin\alpha} > 0.
\end{equation}
Conversely, when $\frac{\cos\alpha}{\sin\alpha} < 0$, indicating that $\alpha$ is an obtuse angle, as illstrated in Fig.~\ref{fig:supp2_figure}~(c), the following relationships hold:
\begin{equation}
	\frac{\cos{\beta}}{\sin{\beta}} = \frac{t+b}{dis_{ref}\sin\beta},
\end{equation}
\begin{equation}
	\small    \frac{\cos{\beta}}{\sin{\beta}}-\frac{t}{dis_{ref}\sin{\beta}} = \frac{b}{dis_{ref}\sin\beta} > 0,~~ \textrm{s.t.}~\frac{\cos\alpha}{\sin\alpha} < 0.
\end{equation}
Combining both cases, we derive the following equation:
\begin{equation}
	\frac{\cos{\alpha}}{\sin{\alpha}}=\frac{t}{dis_{ref}\sin{\beta}} - \frac{\cos{\beta}}{\sin{\beta}},
\end{equation}
and we can rewritten the above equation as follows:
\begin{equation}
	dis_{ref}=\frac{t\sin{\alpha}}{\sin{(\alpha + \beta)}}.\label{equ:19}
\end{equation}
Thus, we obtain the gradient expression of $dis_{ref}$ with respect to $\alpha$, which is formulated as follows:
\begin{equation}
	\frac{\mathrm{d}dis_{ref}}{\mathrm{d}\alpha} = \frac{t\sin{\beta}}{\sin^2{(\alpha + \beta)}}.\label{equ:20}
\end{equation}
\noindent\textbf{Derivation of $\frac{\mathrm{d}\alpha}{\mathrm{d}dis_{pro}}$.} Using a similar approach as for Eq.~(\ref{equ:19}), we can derive the relationship between $\alpha$ and $dis_{pro}$:
\begin{equation}
	dis_{pro} = \frac{m\sin{\alpha}}{\sin{(\alpha + \theta)}}.
\end{equation}
Differentiating both sides with respect to $dis_{pro}$, we obtain $\frac{\mathrm{d}\alpha}{\mathrm{d}dis_{pro}}$, which are formulated as follows:
\begin{equation}
	\small    1 = \frac{m\cos{\alpha}\sin{(\alpha+\theta)}\frac{\mathrm{d}\alpha}{\mathrm{d}dis_{pro}} - m\sin{\alpha}\cos{(\alpha + \theta)\frac{\mathrm{d}\alpha}{\mathrm{d}dis_{pro}}}}{\sin^2{(\alpha + \theta)}},
\end{equation}
\begin{equation}
	\frac{\mathrm{d}\alpha}{\mathrm{d}dis_{pro}} = \frac{\sin^2(\alpha + \theta)}{m\sin{\theta}}.\label{equ:23}
\end{equation}
\noindent\textbf{Final Expression for $dis_{ref}^\prime(dis_{pro})$.} Using Eqs.~(\ref{equ:20}) and (\ref{equ:23}), we obtain the analytical form of $dis_{ref}^\prime(dis_{pro})$, that is
\begin{equation}
	\begin{aligned}
		dis_{ref}^\prime(dis_{pro})&=\frac{\mathrm{d}dis_{ref}}{\mathrm{d}dis_{pro}}\\
		&= \frac{\mathrm{d}dis_{ref}}{\mathrm{d}\alpha}\frac{\mathrm{d}\alpha}{\mathrm{d}dis_{pro}} \\
		&= \frac{t\sin{\beta}\sin^2{(\alpha + \theta)}}{m\sin{\theta}\sin^2{(\alpha + \beta)}}.
	\end{aligned}
\end{equation}
Note that the side lengths in Fig.~\ref{fig:supp2_figure}~(a) (\eg, $t$, $m$) can be derived from the known point coordinates and camera poses. The angles $\alpha$, $\beta$, and $\theta$ can then be calculated using the law of cosines.

\section{Implementation Details}
We implement our method using PyTorch 2.0.0 on an RTX 3090. We conduct 30k iterations for training on the LLFF dataset, 10k iterations for DTU and MipNeRF-360 Datasets, and 4k iterations for Blender dataset. During training, the learning rate for scale was set to 0.03 across the LLFF, MipNeRF-360, and Blender datasets, while the other parameters remain consistent with those used in 3DGS.

\subsection{Dataset Split}
\noindent\textbf{LLFF\&MipNeRF-360}. Following previous methods~\cite{niemeyer2022regnerf,zhu2023fsgs}, we sample images at intervals of 8 from the LLFF~\cite{mildenhall2019local} and MipNeRF-360~\cite{barron2022mip} datasets to create the test set, while the remaining images are used as the training set. For the sparse-view synthesis task, we perform uniform sampling within the training set to select the training views. Consistent with previous work, we downsample all images by a factor of 8.

\noindent\textbf{DTU}. Following previous methods~\cite{yang2023freenerf}, we select 15 scenes from the DTU~\cite{jensen2014large} dataset out of 124, specifically scene IDs 8, 21, 30, 31, 34, 38, 40, 41, 45, 55, 63, 82, 103, 110, and 114. For each scene, views 25, 22, and 28 are used as the 3-view training set, while views 1, 2, 9, 10, 11, 12, 14, 15, 23, 24, 26, 27, 29, 30, 31, 32, 33, 34, 35, 41, 42, 43, 45, 46, and 47 are designated as the test set. All images are downsampled by a factor of 4.

\noindent\textbf{Blender}. In the Blender~\cite{mildenhall2021nerf} dataset, following previous methods~\cite{jain2021putting}, we select views 26, 86, 2, 55, 75, 93, 16, 73, and 8 for training. For evaluation, we uniformly sample 25 images from the test set. All images are downsampled by a factor of 2.

\subsection{Additional Training Details}
During training, we maintain most parameters consistent with those used in 3DGS. Here, we provide additional details beyond those in the main paper. Specifically, the threshold $\epsilon_d$ used in Flow-Filtered Depth Pruning is set to 1.0 for LLFF and DTU, 0.1 for MipNeRF-360, and 0.01 for Blender. The hyperparameter $\lambda_c$ in the objective function was fixed at 0.2. Camera poses are estimated using COLMAP~\cite{schonberger2016structure}, following the methodology of existing sparse-view synthesis studies~\cite{wang2023sparsenerf,li2024dngaussian,zhu2023fsgs}. We utilize FlowFormer++~\cite{shi2023flowformer++} as the optical flow estimator.

\section{Extended Ablation Analysis}
\subsection{Quantitative Ablation Study on DTU Dataset}
\begin{table}
	\centering
	\resizebox{1.0\linewidth}{!}{	
		\begin{tabular}{@{}c|cc|ccc}
			\toprule
			& \multicolumn{2}{c|}{Setting} & PSNR $\uparrow$ & SSIM $\uparrow$  & LPIPS $\downarrow$   \\
			\midrule
			3DGS  & & & 17.65 & 0.816 & 0.146 \\ \midrule
			\multirow{8}{*}{\textbf{Ours}}       & \multicolumn{2}{l|}{\textbf{A}. Average} & 18.92 & 0.836 & 0.128 \\
			& \multicolumn{2}{l|}{\textbf{B}. Nearest} & 19.57 & 0.860 & 0.108 \\
			& \multicolumn{2}{l|}{\textbf{C}. Weighted} & 19.05 & 0.850 & 0.119 \\
			& \multicolumn{2}{l|}{\textbf{D}. FRDB} & 19.76 & 0.864 & 0.109 \\
			& \multicolumn{2}{l|}{\textbf{E}. Average + FFDP} & 18.72 & 0.830 & 0.135 \\
			& \multicolumn{2}{l|}{\textbf{F}. Nearest + FFDP} & 19.74 & 0.860 & 0.108 \\
			& \multicolumn{2}{l|}{\textbf{G}. Weighted + FFDP} & 19.05 & 0.853 & 0.116 \\
			& \multicolumn{2}{l|}{\textbf{H}. FRDB + FFDP} & \textbf{20.21} & \textbf{0.869} & \textbf{0.102}  \\
			\bottomrule
		\end{tabular}
	}
	\vspace{-2mm}\caption{Ablation study on DTU with 3 input views.}
	\label{tab:dtu_ablation}
\end{table}

To complement the ablation results presented in the main paper for the LLFF real-world benchmark, we conduct additional experiments on the object-centric DTU dataset, with the results summarized in Tab.~\ref{tab:dtu_ablation}. Our Flow-Resilient Depth Blending (FRDB) method significantly improves sparse view synthesis performance over variants with alternative blending strategies. Furthermore, when combined with Flow-Filtered Depth Pruning (FFDP), our approach generally outperforms most configurations. This improvement is driven by cleaner, more accurate, and more comprehensive point clouds generated using epipolar depth priors, which lead to enhanced geometric precision and higher-quality details. However, under the Average settings (\textbf{A.} vs. \textbf{E.}), where depth estimates are extremely inaccurate, excessive splitting and replication of erroneous points during refinement lead to performance degradation. By contrast, in scenarios with relatively accurate depth priors, FFDP effectively enhances reconstruction quality by refining point clouds and preserving finer details. Theses consistent performance gains across different datasets demonstrate the robustness and effectiveness of our proposed method.

\subsection{Influence of Distance Threshold}
\begin{table}[t]
		\resizebox{0.47\textwidth}{!}{
			\begin{tabular}{l|cccc}
				\toprule
				Method & Point Number & PSNR $\uparrow$        & SSIM $\uparrow$        & LPIPS $\downarrow$ \\ \midrule
				DNGaussian & 43K & 18.86    & 0.598    & 0.297 \\
				DNGaussian* & 77K & 19.96    & 0.684    & 0.232 \\
				CoR-GS & 80K & 20.29 & 0.705 & 0.201 \\
				FSGS & 299K & 20.34    & 0.695    & 0.207 \\
				\midrule
				Threshold & Point Number & PSNR $\uparrow$        & SSIM $\uparrow$        & LPIPS $\downarrow$ \\ \midrule
				0.01 & 110K & 20.21    & 0.678    & 0.218       \\
				0.1 & 191K & 20.70    & 0.716    & 0.193       \\
				1.0 & 427K & 21.07    & 0.738    & 0.177     \\
				2.0 & 456K & 20.97    & 0.733    & 0.179     \\
				3.0 & 465K & 20.89    & 0.733    & 0.180       \\
				4.0 & 469K & 20.86    & 0.733    & 0.180       \\
				\bottomrule
			\end{tabular}
		}		
	\caption{Influence of distance threshold choices and point cloud comparison with state-of-the-art 3DGS-based competitors on the LLFF dataset with 3 input views. $*$ denotes fused stereo initial point clouds.}\vspace{-2mm}
	\label{table:threshold}
\end{table}
\begin{table}[t]
		\resizebox{1.0\linewidth}{!}{
			\begin{tabular}{l|ccccc}
				\toprule
				Model Type & chairs & kitti & sintel & things\_288960 & things \\ \midrule
				PSNR $\uparrow$ & 21.049    & 21.060    & 21.045  & 21.068  & 21.075 \\
				SSIM $\uparrow$ & 0.738    & 0.738    & 0.738  & 0.739  & 0.738   \\
				LPIPS $\downarrow$ & 0.178    & 0.178    & 0.178  & 0.177  & 0.177   \\
				\bottomrule
			\end{tabular}
		}	
	\caption{The influence of different pre-trained flow estimation models on the LLFF dataset with 3 input views.}\vspace{-2mm}
	\label{tab:flow_psnr}
\end{table}
\begin{table*}[!t]
	\centering
		\begin{tabular}{@{}l|ccc|ccc|ccc}
			\toprule
			\multirow{2}{*}{Method}     & \multicolumn{3}{c|}{2 Views}  & \multicolumn{3}{c|}{3 Views}   & \multicolumn{3}{c}{4 Views}  \\
			&   PSNR$\uparrow$ & SSIM$\uparrow$ & LPIPS$\downarrow$ & PSNR$\uparrow$ & SSIM$\uparrow$ & LPIPS$\downarrow$ & PSNR$\uparrow$ & SSIM$\uparrow$ & LPIPS$\downarrow$ \\
			\midrule		
			RegNeRF                  & 16.16 & 0.396 & 0.455 & 19.08 & 0.587 & 0.336 & 20.95 & 0.689 & 0.272 \\
			FreeNeRF              & \cellcolor{top3}17.12 & \cellcolor{top2}0.490 & \cellcolor{top2}0.364 & 19.63 & 0.612 & 0.308 & \cellcolor{top3} 21.63 & 0.709 & 0.253 \\
			SparseNeRF              & \cellcolor{top2}17.51 & 0.450 & 0.423 & 19.86 & 0.624 & 0.328 & 21.09 & 0.681 & 0.295 \\ \midrule
			3DGS                     & 12.21 & 0.282 & 0.501 & 18.54 & 0.588 & 0.272 & 16.98 & 0.563 & 0.313 \\
			DNGaussian                 & 15.92 & \cellcolor{top3}0.454 & 0.391 & 19.12 & 0.591 & 0.294 & 20.58 & 0.688 & 0.253 \\
			FSGS                     & 16.09 & 0.438 & \cellcolor{top3}0.384 & \cellcolor{top3}20.43 & \cellcolor{top3}0.682 & \cellcolor{top3}0.248 & \cellcolor{top2}21.93 & \cellcolor{top3}0.760 & \cellcolor{top3}0.167 \\
			CoR-GS                  & 14.63 & 0.417 & 0.423 & \cellcolor{top2}20.45 & \cellcolor{top2}0.712 & \cellcolor{top2}0.196 & 21.62 & \cellcolor{top2}0.761 & \cellcolor{top2}0.163 \\
			\textbf{Ours}            & \cellcolor{top1}19.28 & \cellcolor{top1}0.659 & \cellcolor{top1}0.220 & \cellcolor{top1}21.07 & \cellcolor{top1}0.738 & \cellcolor{top1}0.177 & \cellcolor{top1}22.12 & \cellcolor{top1}0.774 & \cellcolor{top1}0.158 \\		
			\bottomrule
		\end{tabular}		
	\caption{Quantitative evaluation of the impact of training views on the LLFF dataset.}
	\label{tab:llff_views}
\end{table*}

We further investigate the impact of varying threshold $\epsilon_d$ in FFDP, as detailed in Tab.~\ref{table:threshold}. The table presents both quantitative metrics and the number of points in the Gaussian representation point cloud. Unlike Tab.~1 in the main paper, where competing methods' results are taken from their original publications, we obtain these metrics by training their official implementations with default settings, as the point counts were not reported.

Examining the lower half of Tab.~\ref{table:threshold}, we observe that when $\epsilon_d$ is set to low values (\eg, 0.1 or 0.01), the performance of NexusGS declines due to the insufficient number of points in the initial point clouds. This reduction in point density leads to excessive splitting in 3DGS, which introduces randomness in point placement. Moreover, the lack of supervision from sparse views prevents the generation of a dense, comprehensive point cloud, ultimately degrading performance.

The optimal results are achieved with a threshold of 1.0. As the threshold increases, more comprehensive and still relatively accurate initial points are obtained, significantly improving the quality of the generated point cloud and the final reconstruction. However, further increases may introduce additional inaccurate points, negatively affecting performance. Despite this, we find that as the threshold grows beyond 1.0, the impact on PSNR becomes more noticeable, while SSIM and LPIPS—metrics that align better with human visual perception—remain less affected. This suggests that our method exhibits tolerance for erroneous initial points, maintaining stable performance while revealing rich high-frequency details in the output.

We also quantitatively analyze the quality of the point clouds generated by state-of-the-art methods such as DNGaussian, FSGS, and CoR-GS, as shown in the upper half of Tab.~\ref{table:threshold}. Regardless of whether random or fused stereo initial point clouds (indicated by an asterisk) are used, DNGaussian, as a lightweight design-focused method, consistently generates fewer points and provides less comprehensive coverage than our approach, even at $\epsilon_d = 0.01$, resulting in inferior performance. Although FSGS generates more points, its limited point addition strategy results in lower accuracy and coverage. Notably, even with a reduced number of points (\eg, $\epsilon_d = 0.1$), our method still outperforms FSGS. As for CoR-GS, while it generates relatively comprehensive coverage, it lacks the ability to produce a dense point cloud. This limitation is reflected in the point count, ultimately restricting the reconstruction quality, especially in high-frequency details. In contrast, NexusGS, with epipolar depth priors, generates a more accurate, dense, and comprehensive point cloud, leading to superior reconstruction performance.

\subsection{Robustness Across Various Flow Estimators}
Inspired by existing approaches~\cite{li2024dngaussian, zhu2023fsgs}, which utilize monocular depth estimators to provide depth priors, we hypothesize that similar variability in performance might occur when using different optical flow estimators with varying network parameters. To explore this possibility, we conduct experiments on the LLFF dataset, comparing the performance of our method using optical flow estimators trained on five different datasets. Specifically, we evaluate the $\textit{chairs}$, $\textit{kitti}$, $\textit{sintel}$, $\textit{things\_288960}$, and $\textit{things}$ models of FlowFormer++. The quantitative results are summarized in Tab.~\ref{tab:flow_psnr}. As illustrated, despite utilizing different pretrained flow estimation models, our method consistently shows minimal variations in PSNR, with SSIM and LPIPS scores remaining nearly identical across the different models. These results highlight the robustness of our approach, demonstrating its effectiveness regardless of the specific optical flow estimator employed.

\subsection{Robustness on Varying View Counts}
To validate the robustness of our method under varying numbers of training views, we conduct experiments on the LLFF dataset, with results presented in Tab.~\ref{tab:llff_views}. Our method consistently outperforms all competitors when using 2, 3, and 4 views. Notably, with only 2 training views, COLMAP fails to generate a fused stereo point cloud, leading to poor performance by DNGaussian, FSGS, and CoR-GS, which perform even worse than NeRF-based methods. In contrast, our method does not suffer from this limitation. By leveraging the point cloud generated through our approach, we achieve superior results with just 2 views, effectively overcoming the constraints of previous 3DGS methods and surpassing NeRF-based methods across all evaluated metrics.

\begin{figure*}
	\centering
	\includegraphics[width=\textwidth]{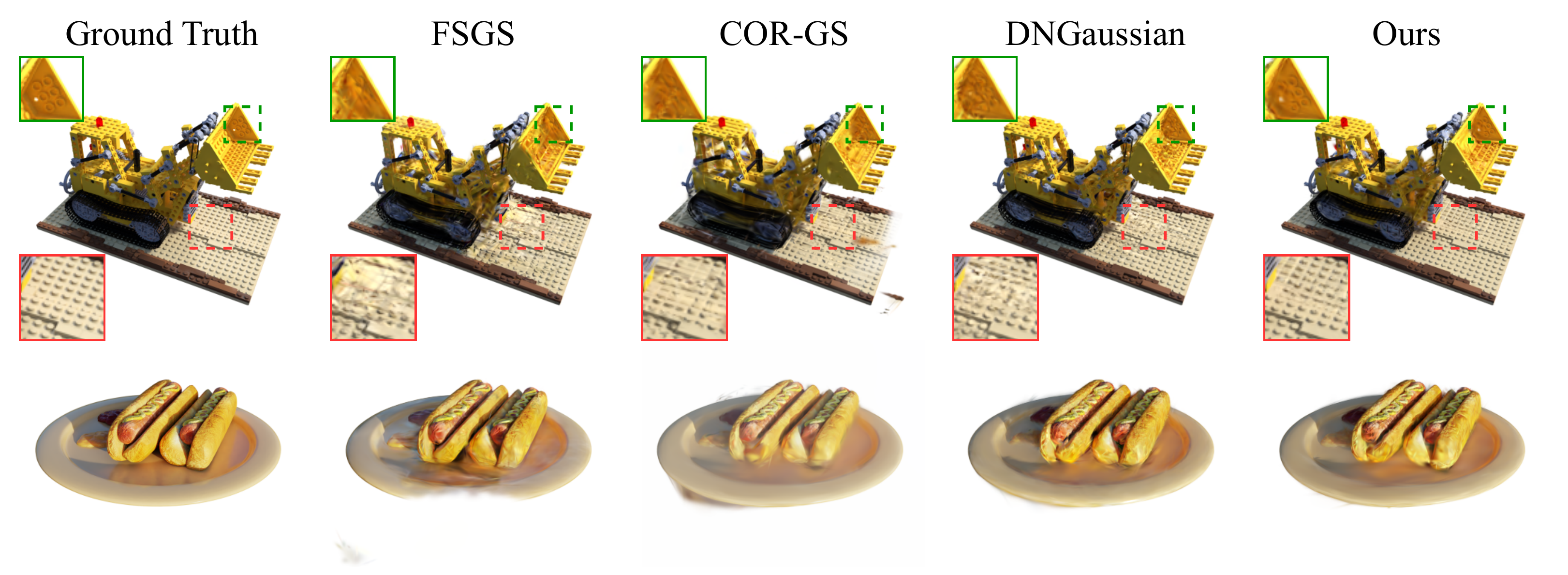}
	\caption{Visual results on the Blender dataset with 8 input views.}\label{fig:blender}
\end{figure*}
\begin{figure*}[!t]
	\setlength{\abovecaptionskip}{12pt}
	\centering
	\includegraphics[width=1\linewidth]{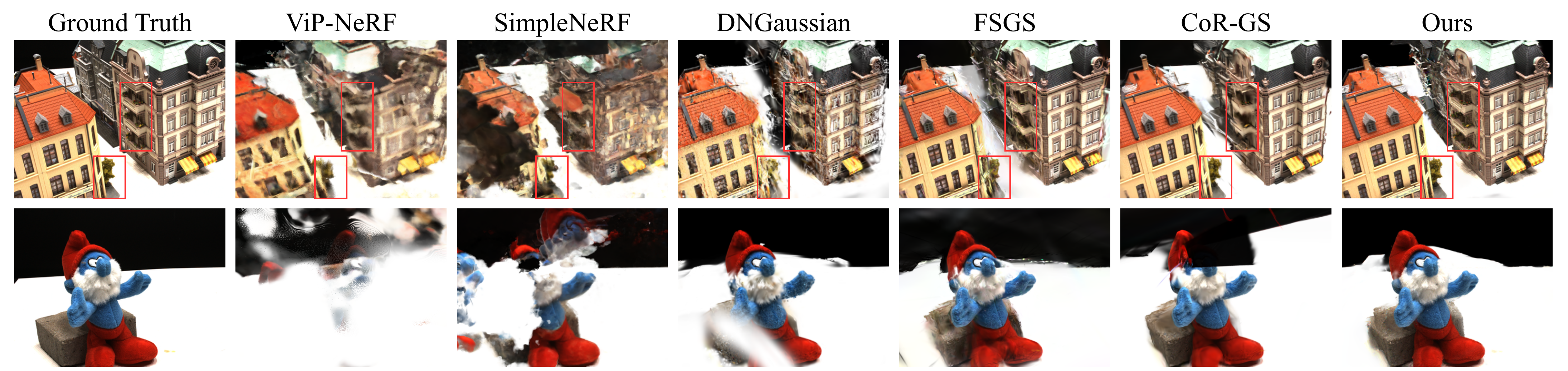}
	\caption{
		Additional visual comparisons on the DTU dataset with 3 input views.
	}
	\label{fig:dtu_supp_result}
\end{figure*}

\section{Additional Visual Results}
	\begin{figure*}[htp]
		\centering
		\includegraphics[width=0.95\textwidth]{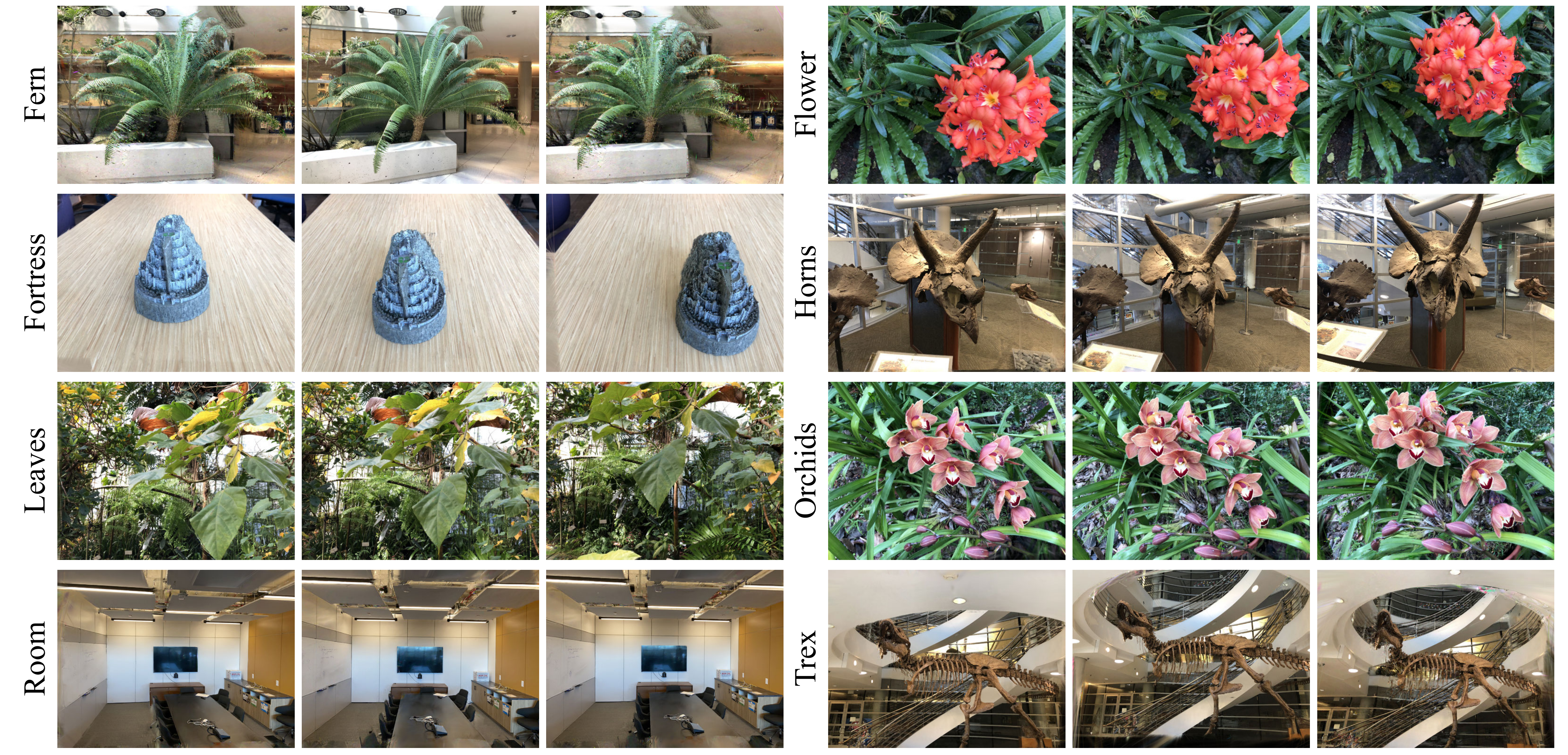}
		\caption{Per-scene visual results on the LLFF dataset with 3 input views.}\label{fig:LLFF}
	\end{figure*}
	
	\begin{figure*}[htp]
		\centering
		\includegraphics[width=0.95\textwidth]{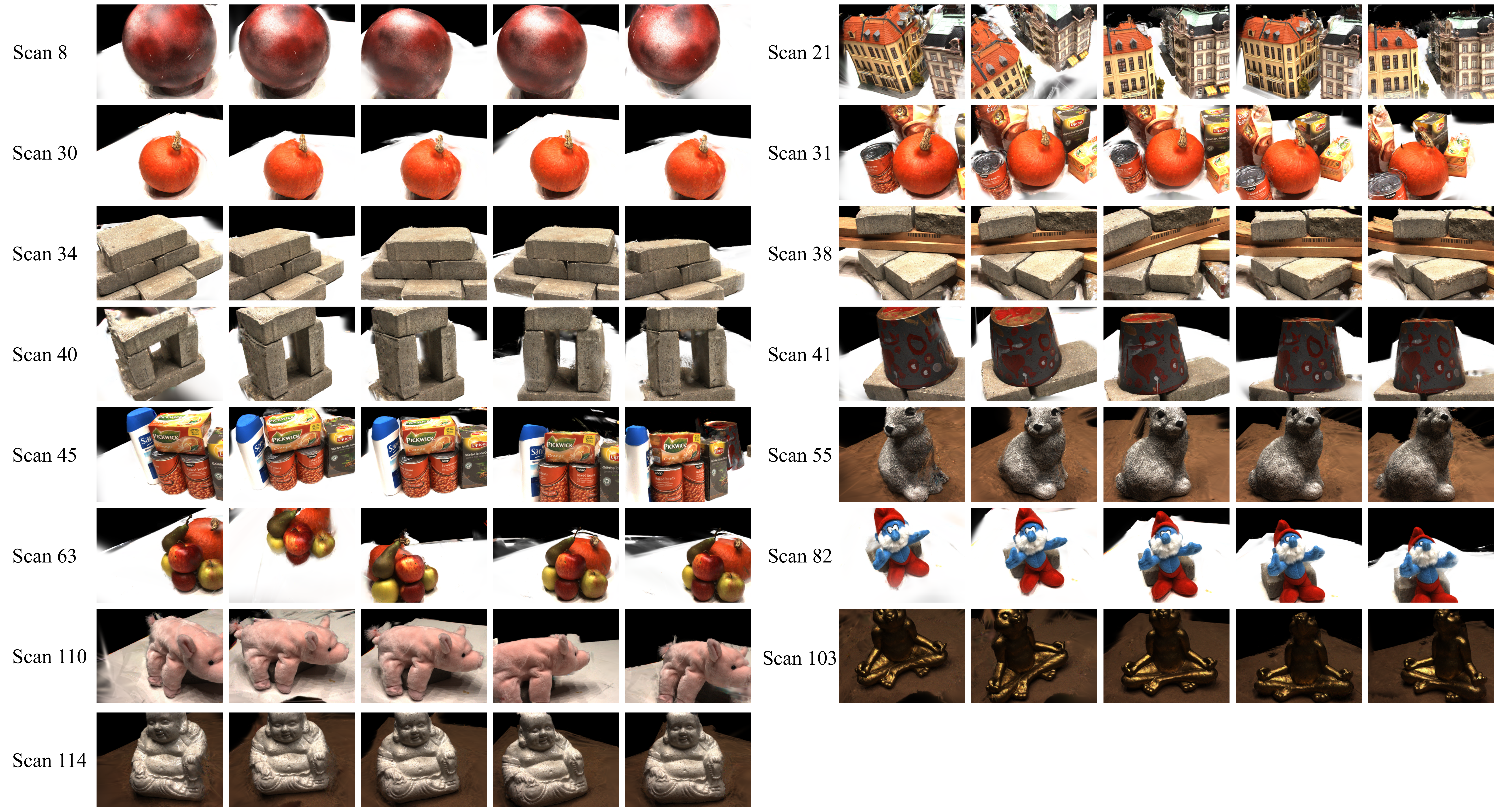}
		\caption{Per-scene visual results on the DTU dataset with 3 input views.}\label{fig:DTU}
	\end{figure*}
Additional visual results are provided in the supplementary materials. Specifically, Fig.~\ref{fig:blender} showcases the results of our method on the Blender dataset. As shown, our method achieves a more complete and detailed reconstruction compared to previous approaches. This improvement is attributed to the accurate geometry provided by the precise point cloud, which enables the faithful reconstruction of finer textures.
	
We also present additional visual results on the DTU dataset in Fig.~\ref{fig:dtu_supp_result}. It is evident that the results from NeRF-based methods, constrained by depth priors, are adversely affected by floaters, leading to degraded reconstruction quality. This suggests that their depth priors are not well-suited for object-centered datasets. In contrast, subsequent 3DGS-based methods that utilize depth priors mitigate the presence of floaters. However, due to the lack of dense and accurate point clouds, their final reconstructions appear visually smoother, lacking finer detail.
	
	\begin{figure*}[htp]
		\centering
		\includegraphics[width=1.0\textwidth]{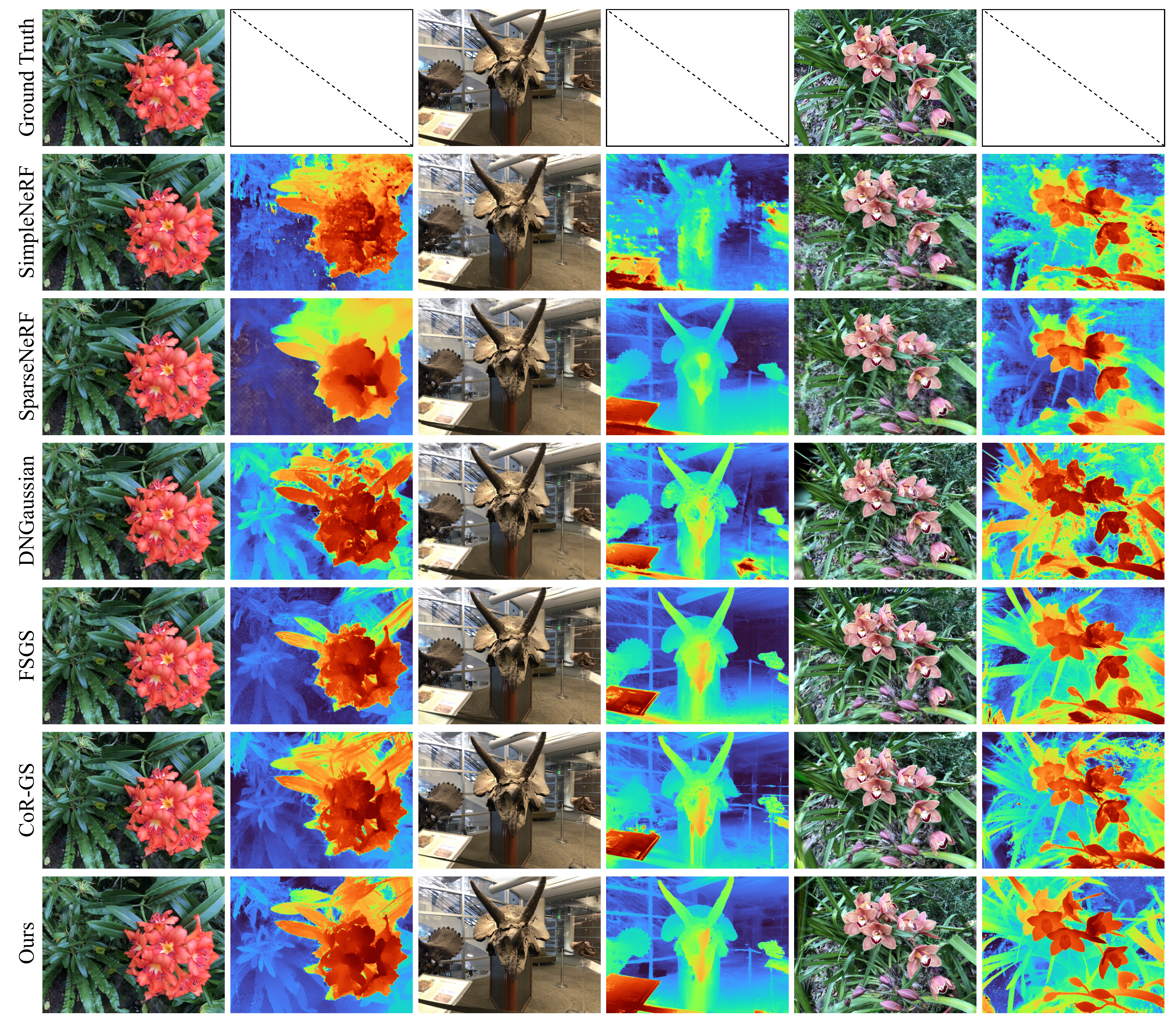}
		\caption{Visual comparisons of depth maps on the LLFF dataset with 3 input views.}\label{fig:depth}
	\end{figure*}
	
Furthermore, Figs.~\ref{fig:LLFF} and \ref{fig:DTU} present the results for each test scene on the LLFF and DTU datasets, respectively. Fig.~\ref{fig:depth} compares the depth maps of our method and the competitors.

\section{Discussion, Limitations, and Future Work} 
Despite NexusGS's remarkable performance and generalizability, it, like most sparse-view synthesis methods, relies on known camera poses to enforce epipolar constraints. To evaluate the robustness of our method, we conduct a sensitivity analysis of camera calibration errors, using the SSIM score as the primary metric, as summarized in Tab.~\ref{sen}. Perturbations are introduced to the camera pose along all axes, resulting in performance degradation across all competitors, all of which depend on accurate pose information. Severe overfitting is observed when perturbations reach 0.1. In contrast, our method demonstrates superior robustness, consistently outperforming others despite pose errors, thanks to our effective error-handling strategies, FRDB and FFDP.
	
	\begin{table}
		\centering	
		\resizebox{0.47\textwidth}{!}{
			\begin{tabular}{l|c|c|c|c|c|c}
				\toprule
				Perturbation & 0 & 0.02 & 0.04 & 0.06 & 0.08 & 0.1 \\ 
				\midrule
				DNGaussian & 0.591 & 0.586  & 0.565  & 0.556  & 0.540 & 0.476  \\
				FSGS &  0.682  & 0.688   & 0.664  & 0.633  & 0.606 & 0.579  \\
				CoR-GS & 0.712  & 0.693   & 0.664  & 0.639  & 0.609 & 0.586  \\
				\midrule
				Ours &  \textbf{0.738}  & \textbf{0.724}   & \textbf{0.690}  & \textbf{0.653}  & \textbf{0.617} & \textbf{0.588}  \\		
				\bottomrule
		\end{tabular}}
		\caption{Sensitivity analysis of calibration errors on the LLFF dataset with 3 input views using SSIM.}\label{sen}\vspace{-4mm}
	\end{table}
	
Although recent pose-free methods, such as COGS~\cite{jiang2024construct}, bypass the need for camera poses, the trade-off between flexibility and rendering accuracy leaves room for improvement, offering a promising avenue for future work.

%%%%%%%%% REFERENCES
{\small
\bibliographystyle{ieee_fullname}
\bibliography{cvpr25}
}

\end{document}